\documentclass[11pt]{article}

\usepackage[preprint]{acl}

\usepackage{times}
\usepackage{latexsym}

\usepackage[T1]{fontenc}
\usepackage[utf8]{inputenc}

\usepackage{microtype}

\usepackage{inconsolata}

\usepackage{graphicx}

\usepackage{hyperref}       % hyperlinks
\usepackage{url}            % simple URL typesetting
\usepackage{booktabs}       % professional-quality tables
\usepackage{amsfonts}       % blackboard math symbols
\usepackage{nicefrac}       % compact symbols for 1/2, etc.
\usepackage{microtype}      % microtypography
\usepackage{xcolor}         % colors
\usepackage{amsthm}        % theorem environments

\usepackage{amsmath, amssymb, amsthm, bm}
\usepackage{mathtools}
\usepackage{tabularx}
\usepackage{multirow}
\usepackage{algorithm}
\usepackage{algpseudocode}
\usepackage{tikz}
\usetikzlibrary{arrows.meta, positioning, shapes.geometric}
\usepackage{graphicx}
\usepackage{float}
\usepackage{subcaption}
\usepackage{enumitem}
\usepackage{natbib}
\DeclareMathOperator*{\argmin}{arg\,min}
\DeclareMathOperator*{\argmax}{arg\,max}
\newcommand{\hvec}{\bm{h}}
\newcommand{\Rmat}{\bm{R}}
\newcommand{\Wmat}{\bm{W}}
\newcommand{\bvec}{\bm{b}}
\newcommand{\phiparam}{\phi}

\newcommand{\KLdiv}{\mathrm{KL}}

\newcommand{\Dt}{\mathcal{D}_{t}}

\newtheorem{proposition}{Proposition}

\definecolor{myblue}{RGB}{31, 119, 180}
\definecolor{mygreen}{RGB}{60,140,90}
\usetikzlibrary{arrows.meta, positioning}
\definecolor{myorange}{RGB}{230,140,60}
\definecolor{myred}{RGB}{200,70,70}

\title{CRAFT: Forgetting-Aware Intervention-Based Adaptation for Continual Learning}
\author{
  Md Anwar Hossen$^{1}$ \thanks{Corresponding author.} \quad
  Fatema Siddika$^{1}$\ \quad
  Juan Pablo Mu\~noz$^{2}$ \\
  Tanya Roosta$^{3,4}$ \quad
  Ali Jannesari$^{1}$ \\[4pt]
  $^{1}$Iowa State University,\quad
  $^{2}$Maro Systems, USA\quad\\
  $^{3}$University of California, Berkeley,\quad$^{4}$AMD \\[2pt]
  \texttt{\{manwar, fatemask, jannesar\}@iastate.edu} \\
  \texttt{pablo.munoz@maro-systems.com,}\quad\texttt{ tanya.roosta@gmail.com}
}

\begin{document}
\maketitle
\begin{abstract}
Large language models (LLMs) can learn new tasks through fine-tuning, but learning tasks sequentially often leads to catastrophic forgetting of earlier ones. This paper proposes CRAFT, a continual learning framework that keeps model weights frozen and instead learns low-rank interventions on hidden representations. Similar tasks share an intervention to help knowledge transfer, with similarity judged by how each task alters the model's output distribution rather than by a learned gating network. Within a shared intervention, a Kullback–Leibler (KL) penalty keeps each new task close to the intervention's existing behavior, which limits forgetting of earlier tasks, while a small, task-private intervention in the deepest layers preserves dedicated capacity that later tasks cannot overwrite. The learned interventions also make it possible to identify the task at test time, so task identity labels are not needed. Across multiple benchmarks and model scales, CRAFT improves overall performance and reduces forgetting compared with strong LoRA-based methods, while remaining robust to task ordering.
Code Repository: \url{https://github.com/mahemon/CRAFT-CL}
\end{abstract}
\section{Introduction}
\label{sec:intro}
Large language models are not designed to learn continuously. Each new task updates the same parameters that encode prior knowledge, making catastrophic forgetting an inherent consequence of fine-tuning \citep{vela2022temporal}. Existing approaches attempt to manage this interference through regularization \citep{kirkpatrick2017overcoming}, replay~\citep{shin2017continual}, or architectural constraints, but all operate within parameter space, where the conflict itself originates.

This paper takes a different approach and moves continual learning out of parameter space. Representation fine-tuning~\citep{wu2404reft} edits hidden states within a low-rank subspace while the backbone stays frozen. Because the backbone never changes, weight-level forgetting is avoided. Forgetting is not eliminated, however. When tasks share an intervention, interference can still arise within it. What changes is where interference lives. It leaves the model weights and is confined to a low-dimensional intervention subspace, where it can be directly measured and controlled. Continual learning thus shifts from a problem of parameter competition to one of organizing, sharing, and updating representation-level interventions.

This reframing raises three questions. First, \textit{where should adaptation reside?} Isolated interventions prevent transfer, while a single shared intervention forces unrelated tasks to compete. Second, \textit{how should tasks be assigned during training without a learned router}, whose gate adds parameters that can themselves forget? Third, \textit{how can forgetting be controlled without limiting learning?} To our knowledge, these questions have not been addressed jointly within a representation fine-tuning framework.

This paper presents CRAFT, Continual Representation-Anchored Fine-Tuning, a framework built on representation fine-tuning. CRAFT handles each new task in three steps. First, a brief warm-up produces an output-distribution signature of the task, and a KL distance on output distributions routes it to the shared intervention of similar tasks or opens a new one. Second, the task is fine-tuned with a KL penalty toward the shared intervention's current state, a quantity linked to forgetting in prior work~\citep{shenfeld2026rl}, so it learns the new task while limiting drift on earlier tasks. Third, the trained intervention is merged back into the shared intervention with a weight set by the same KL signal. A small part of each task's intervention stays private to the task. At test time, when task identity is unavailable, CRAFT identifies the task from hidden states read after each learned shared intervention. The main contributions of this work are as follows.
\begin{itemize}[leftmargin=*, itemsep=1pt, topsep=3pt]
\item CRAFT brings representation fine-tuning to continual learning in large language models and, to our knowledge, is the first framework that learns a task sequence through low-rank interventions on the hidden representations of a frozen backbone.
\item During training, each new task is routed to a shared intervention of similar tasks without a learned gate, letting related tasks share knowledge. At test time, the task is identified from the learned interventions rather than the unedited backbone, and this routing does not forget earlier tasks.
\item A single KL quantity both limits drift during training and sets how much of the trained state is merged back into the shared intervention, so a task that drifts far is committed with a small weight.
\item On TRACE and a 15-task long sequence with three base models, CRAFT improves overall performance over the strongest baselines while reducing forgetting.
\end{itemize}
\section{Related Work}
\label{sec:related}
Continual learning for LLMs is commonly evaluated on TRACE~\citep{wang2023trace}, which chains classification, summarization, code and reasoning tasks. Existing methods act in parameter space. Regularization methods such as EWC~\citep{kirkpatrick2017overcoming} and SI~\citep{zenke2017continual} penalize changes to weights that matter for earlier tasks, and replay methods such as Self-Synthesized Rehearsal~\citep{huang2024mitigating} tune the weights on generated examples of past tasks. Adapter methods isolate tasks instead: O-LoRA~\citep{wang2023orthogonal} and InfLoRA~\citep{liang2024inflora} restrict each task to a low-interference subspace, TreeLoRA~\citep{qian2025tree} groups adapters by gradient similarity, and mixture-of-LoRA methods such as MoELoRA~\citep{liu2023moelora}, D-MoLE~\citep{ge2025dynamic} and GainLoRA~\citep{liang2025gainlora} train gates jointly with the adapters. SETA~\citep{siddika2026seta} learns each task in sparse expert blocks of a frozen backbone, shared across tasks by position, in a centralized setting. \citet{anonymous2026fedseam} extends sparse expert adaptation to federated continual learning, aggregating clients' experts on a server. LoReFT~\citep{wu2404reft} instead edits hidden states in a low-rank subspace of a frozen backbone and matches weight-space fine-tuning with fewer parameters, but it has been studied only on single tasks in language models. KL penalties against an anchor distribution are standard in RLHF~\citep{stiennon2020learning, ouyang2022training}, and recent work links lower forgetting to a small KL divergence on the new task~\citep{chen2025retaining, shenfeld2026rl}, always in parameter space and against the base model. ~\ref{sec:litreview} discusses each related work in more detail.
\section{Problem formulation}
\label{sec:problem}
\subsection{Beyond weight space: adapting representations}
CRAFT builds on Low-rank Linear Subspace ReFT (LoReFT)~\citep{wu2404reft}, which edits the residual stream, the hidden states $\hvec^{(l)} \in \mathbb{R}^d$ passed between Transformer layers $l=1,\dots,L$. An intervention at layer $l$ modifies $\hvec^{(l)}$ at selected token positions. LoReFT edits a hidden state as
\begin{equation}
  \Phi(\hvec^{(l)};\phiparam) = \hvec^{(l)} + \Rmat^{\top}(\Wmat \hvec^{(l)} + \bvec - \Rmat \hvec^{(l)}),
  \label{eq:loreft}
\end{equation}
where $\phiparam = \{\Rmat, \Wmat, \bvec\}$, $\Rmat \in \mathbb{R}^{r \times d}$ has orthonormal rows with $r \ll d$, and $\Wmat \in \mathbb{R}^{r \times d}$ and $\bvec \in \mathbb{R}^{r}$ define the learned edit within that subspace. Only this $r$-dimensional subspace of the representation changes, and the rest stays intact.
\begin{figure*}[t]
\centering
\includegraphics[width=1\linewidth]{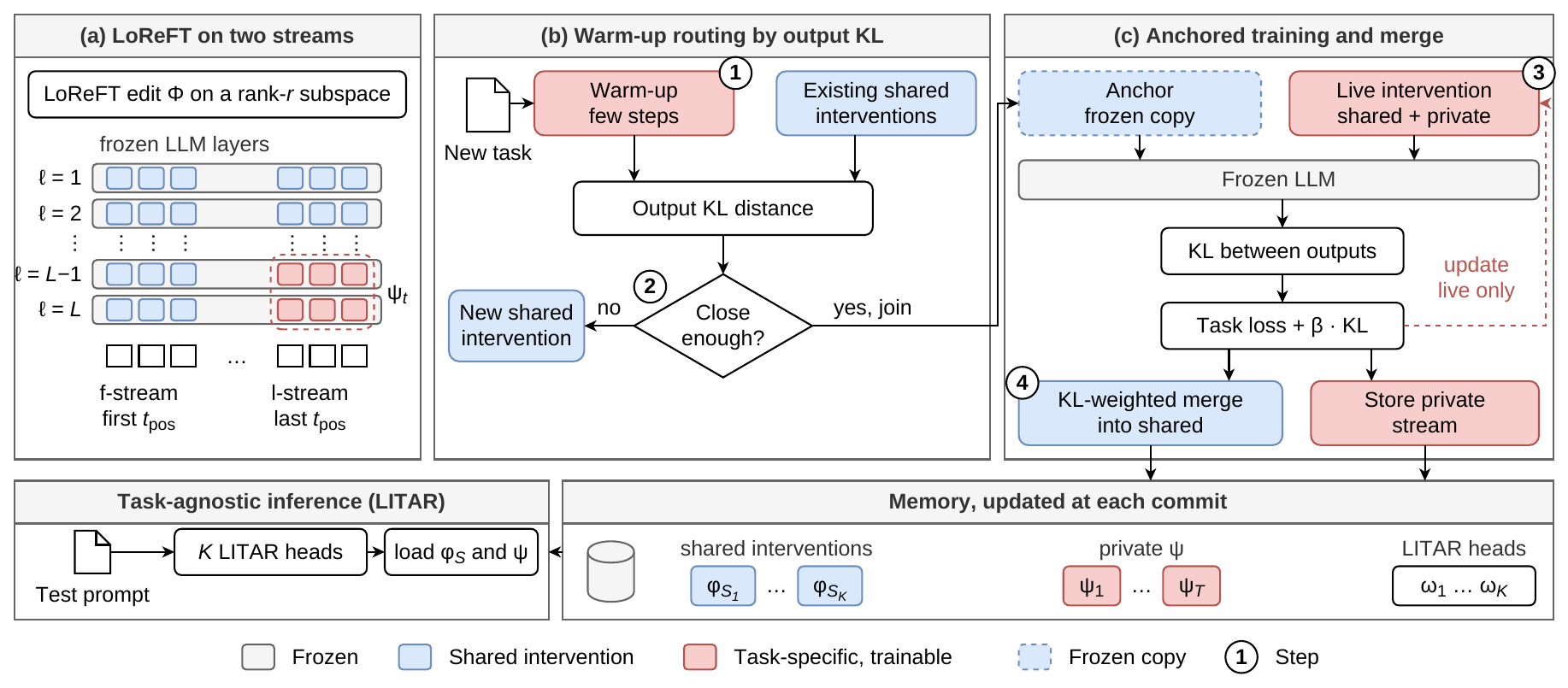}
\caption{Overview of CRAFT. (a) Interventions edit the first and last prompt tokens of a frozen LLM, and the deepest layers keep a small part private to each task. (b) After a brief warm-up, a new task joins the shared intervention whose outputs it would change least, or opens a new one if all are too far. (c) The task trains with a KL penalty toward that shared intervention, is merged back with a KL-dependent weight, and keeps its private part. At test time, task-agnostic routing identifies the task from the learned interventions and loads its state.}
\label{fig:arch}
\end{figure*}
\subsection{Continual setting and research questions}
Tasks $\tau_1,\dots,\tau_T$ arrive in sequence, each with training data $\mathcal{D}_t$, while the backbone weights $\theta$ stay frozen. CRAFT keeps one shared intervention state $\phiparam_{\mathcal{S}_k}$ for each set $\mathcal{S}_k$ of similar tasks. The problem is to decide which set a new task joins and to update $\phiparam_{\mathcal{S}_k}$ for that task without degrading the tasks it already holds. Three research questions follow from this setting. \\
\textbf{Unit of intervention adaptation:}
LoReFT was developed for single tasks, so it does not say how a task stream should be divided across interventions. \textbf{RQ1:} Can interventions be shared across related tasks while staying isolated across tasks that would interfere? \textbf{C1:} Intervention-level continual learning with stream separation. Interventions sit on two prompt streams. The f-stream covers the first $t_{pos}$ tokens, where instructions are encoded, and the l-stream covers the last $t_{pos}$ tokens, where the output is committed. All intervention sites except the l-stream of the deepest layers form the shared state $\phiparam_{\mathcal{S}_k}$, and the remaining sites form a small task-private intervention $\psi_t$. Tasks within $\mathcal{S}_k$ transfer knowledge through $\phiparam_{\mathcal{S}_k}$, and routing keeps tasks in different shared interventions from interfering. \\
\textbf{Task routing on output-distribution divergence:}
Mixture-of-LoRA approaches~\citep{liu2023moelora, yang2024moral, ge2025dynamic} pair a pool of adapters with a learned gate. The gate is itself a set of weights that can forget, and it depends on task identifiers or held-out data, so routing is decided by an auxiliary network rather than by the task representation. \textbf{RQ2:} Can a task be assigned during training from how a brief warm-up would change the outputs the system already produces, without any trainable routing component? \textbf{C2:} Routing on output-distribution divergence. After a warm-up of $S_{\mathrm{wu}}$ steps from a shared random seed, the new task induces an output distribution $\pi_{\widetilde{\phiparam}^{(t)}}$. CRAFT computes its symmetric KL distance to the output distribution of each existing shared intervention on a few stored prompts of its member tasks, normalized by how far each side has moved from the unadapted model. The task joins the closest shared intervention if the distance is below $\delta$ and otherwise opens a new one. \\ 
\paragraph{Learning new tasks with less forgetting:}
Adapting a new task inside a shared intervention risks overwriting what that intervention already encodes. Existing methods anchor updates to fixed references that do not encode past tasks~\citep{yu2024language, yadav2023ties}, distill toward the whole previous model~\citep{li2017learning}, or constrain gradients without deciding how the update is integrated~\citep{chaudhry2018efficient}, so no single signal decides both how far adaptation should go and how much of it is kept. \textbf{RQ3:} Can a new task be learned without overwriting prior knowledge, and which signal should control drift and decide how the update is integrated? \textbf{C3:} KL as a unified training-time signal. Once a task is routed, CRAFT freezes the current state of its shared intervention as an anchor $\phiparam^{\mathrm{anchor}}$ and trains the task's intervention $\phiparam^{(t)}$ with a forward KL penalty
\begin{equation}
  \mathcal{L}_{\mathrm{KL}}
  = \KLdiv\bigl(\pi_{\phiparam^{\mathrm{anchor}}} \,\|\,
                \pi_{\phiparam^{(t)}}\bigr).
  \label{eq:kl-control}
\end{equation}
Prior work shows empirically that this KL predicts how much fine-tuning forgets~\citep{shenfeld2026rl}. The trained shared parameters $\phiparam^{(t)}_{\text{shared}}$ are then interpolated into the shared intervention state with a weight that shrinks as the final KL grows, as in Eq.~\eqref{eq:kl_merge}, and the private part $\psi_t$ is stored for task $t$ alone. The same KL signal therefore controls both drift during training and integration afterward.
\begin{figure*}[t]
\centering
\includegraphics[width=0.75\textwidth]{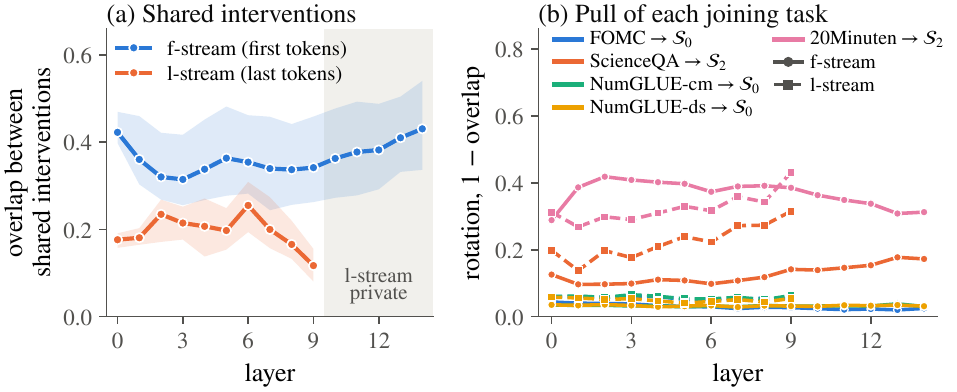}
\caption{Geometry of the shared interventions on TRACE-8 with Llama-3.2-1B-Instruct. (a) Subspace overlap between the final shared interventions at each layer, mean over pairs with the range shaded; the l-stream of the deepest layers is private to each task. (b) Rotation of a shared intervention when a task joins it, measured between its states before and after the merge.}
\label{fig:intervention_geometry}
\end{figure*}
\section{Methodology}
\label{sec:method}
\subsection{Design choice of intervention layers and token streams}
\label{sec:setup} 
CRAFT is designed so that each new task is placed with compatible tasks, learns without disrupting what earlier tasks have learned, and updates the shared knowledge in a controlled way. These goals lead to the three-stage pipeline shown in Figure~\ref{fig:arch}, in which the task is first routed by the KL divergence between output distributions, then trained with a KL penalty, and finally merged with a weight set by the same KL. At every Transformer layer, the LoReFT interventions of Eq.~\eqref{eq:loreft} act on two streams of prompt tokens: the f-stream covers the first $t_{pos}$ positions, and the l-stream covers the last $t_{pos}$ positions. Tasks with similar outputs share one intervention $\phiparam_{\mathcal{S}_{k}}$, which holds every site except the l-stream of the deepest layers, and each task keeps those remaining sites as its own private part $\psi_t$.
\subsection{Task routing on output-distribution divergence}
\label{sec:routing}
CRAFT routes tasks by their output distributions rather than their intervention weights, because outputs are comparable across tasks while weights depend on how each subspace is parameterized. When a new task $t$ arrives, a LoReFT module initialized from a shared random seed is trained for $S_{\mathrm{wu}}$ warm-up steps on $\mathcal{D}_t$, giving $\widetilde{\phiparam}^{(t)}$. Since forgetting concerns the tasks already in a shared intervention, the comparison uses stored probes $\mathcal{P}_s$ of $n_p$ training prompts from each member $s$ of $\mathcal{S}_k$. On $\mathcal{P}_s$, $D^{s}_{K}$ and $D^{s}_{S_k}$ measure how far the new task and the shared intervention have moved from the unadapted model $\pi_0$, and $D^{s}_{KS_k}$ measures how far they are from each other, as defined in Eq.~\eqref{eq:route_div} and shown in Figure~\ref{fig:arch}b. The routing distance
\begin{equation}
\begin{gathered}
  d(t,k) = \frac{\sum_{s \in \mathcal{S}_k} D^{s}_{KS_k}}
                {\max\bigl(\sum_{s \in \mathcal{S}_k} m^{s}_{k},\, \varepsilon_{\mathrm{b}}\bigr)},\\
  m^{s}_{k} = \max\bigl(\min(D^{s}_{K}, D^{s}_{S_k}),\, \varepsilon_{\mathrm{b}}\bigr),
\end{gathered}
\label{eq:route_dist}
\end{equation}
compares this gap with how far each side has moved, so two models that barely moved do not appear close. Task $t$ joins the closest shared intervention if $d(t,k^{\star}) \le \delta$ and its own probe gives $D_{KS_{k^{\star}}} \ge \varepsilon_{\mathrm{sig}}$, and otherwise opens a new shared intervention from the fixed-seed initialization.

\subsection{Anchored training and KL-weighted merge}
\label{sec:phase2}
Once task $t$ joins $\mathcal{S}_k$, its intervention $\phiparam^{(t)}$ is initialized from $\phiparam_{\mathcal{S}_k}$ on the shared sites and from the fixed seed on the private sites. A frozen copy of this starting point serves as the anchor $\phiparam^{\mathrm{anchor}}$ and records what $\mathcal{S}_k$ encoded before $t$ arrived. The anchor is the shared state rather than the unadapted model $\pi_0$, because $\pi_0$ encodes none of the tasks routed to $\mathcal{S}_k$, and pulling toward it would erase what the shared intervention has learned. For each batch, CRAFT minimizes
\begin{equation}
  \mathcal{L}_{\mathrm{total}} = \mathcal{L}_{\tau}(\phiparam^{(t)}) + \beta\,\mathcal{L}_{\mathrm{KL}},
  \label{eq:kl_loss}
\end{equation}
where $\mathcal{L}_{\mathrm{KL}}$ is the forward KL of Eq.~\eqref{eq:kl-control}, averaged over the positions that predict answer tokens and used without normalization, so $\beta$ directly sets its weight. After training, $\psi_t$ is stored and the shared part is merged as
\begin{equation}
  \phiparam_{\mathcal{S}_k} \leftarrow w_{t}\,\phiparam^{(t)}_{\text{shared}} + (1 - w_{t})\,\phiparam_{\mathcal{S}_k},
  \qquad
  w_{t} = \frac{h}{h + \bar{k}_{t}},
  \label{eq:kl_merge}
\end{equation}
where $\bar{k}_{t}$ is the KL to the anchor on the training data of $t$ and $h$ is the KL at which $w_t = 1/2$, so a task that drifted further is merged less. A task that opens a new shared intervention has no members to protect, so it trains on $\mathcal{L}_{\tau}$ alone and its trained shared state becomes $\phiparam_{\mathcal{S}_k}$ without a merge. Tasks in other shared interventions are never touched, and an earlier task in the same shared intervention changes only through this merge.
\subsection{Task-agnostic inference}
\label{sec:eval-path}
CRAFT must identify the task itself while doing inference, because $\psi_t$ belongs to a single task. LoReFT-intervention task-agnostic routing (LITAR) does this with one linear softmax classifier $\omega_k$ per shared intervention, giving $p_{\omega_k}(\tau \mid x)$ over the tasks seen so far. Its features are the hidden states of prompt $x$ with $\phiparam_{\mathcal{S}_k}$ loaded and the private sites unedited, averaged over the last $t_{pos}$ prompt tokens and concatenated across layers, so the classifier reads the representation after the learned intervention. At every commit, CRAFT recomputes the features of the changed shared intervention on all stored prompts, adds the new task's features to the others, and refits every $\omega_k$ from scratch, so no classifier forgets an earlier task. At test time, LITAR averages its predictions over the first $m$ prompts $\mathcal{Q}$ of a test task and over all $K$ shared interventions,
\begin{equation}
  \hat{\tau} = \argmax_{\tau} \tfrac{1}{mK}\textstyle\sum_{x \in \mathcal{Q}}\sum_{k=1}^{K} p_{\omega_k}(\tau \mid x),
  \label{eq:task_head_decision}
\end{equation}
and CRAFT evaluates the task with $\phiparam_{\mathcal{S}_{M[\hat{\tau}]}}$ and $\psi_{\hat{\tau}}$. Training-time routing still uses Eq.~\eqref{eq:route_dist}, so an error affects only the state loaded at test time. ~\ref{app:litar} gives the feature standardization, the classifier fit and the inference cost.
\section{Theoretical analysis}
\label{sec:theory}
This section states what the design of CRAFT guarantees exactly, what the KL-weighted merge controls locally, and what remains an empirical assumption. For a member $s$ of $\mathcal{S}_k$, let $P^{-}_s(\cdot \mid x)$ and $P^{+}_s(\cdot \mid x)$ be the output distributions of its deployed configuration $(\phiparam_{\mathcal{S}_k}, \psi_s)$ before and after the commit of a new task $t$. The commit replaces $\phiparam_{\mathcal{S}_k}$ by $\phiparam_{\mathcal{S}_k} + w_t \bm{\delta}_t$ with $\bm{\delta}_t = \phiparam^{(t)}_{\text{shared}} - \phiparam_{\mathcal{S}_k}$, as in Eq.~\eqref{eq:kl_merge}. The change on $s$ is measured by any per-example score $G \in [0,1]$ of a sampled output, $\Delta_s = \mathbb{E}_{x \sim \mathcal{D}_s}\bigl[\mathbb{E}_{P^{+}_s} G - \mathbb{E}_{P^{-}_s} G\bigr]$.

\paragraph{Isolation across shared interventions.}
Training task $t$ updates only $\phiparam^{(t)}$, and its commit changes only $\phiparam_{\mathcal{S}_{M[t]}}$ and adds $\psi_t$. Under task-aware inference, every task in another shared intervention is evaluated with the same state before and after $t$, so its outputs are unchanged. Within $\mathcal{S}_{M[t]}$, a member can change only through the commit, because no private part $\psi_s$ with $s \neq t$ is ever updated. With task-agnostic routing, the same holds for every task whose predicted identity does not change after the refit.

\begin{proposition}[Forgetting is bounded by the member's own KL]
\label{prop:member-kl}
For every member $s$ and every score $G \in [0,1]$,
\begin{equation}
  |\Delta_s| \le \sqrt{\tfrac{1}{2}\,\mathbb{E}_{x \sim \mathcal{D}_s}\KLdiv\bigl(P^{-}_s(\cdot \mid x) \,\|\, P^{+}_s(\cdot \mid x)\bigr)},
  \label{eq:member-kl}
\end{equation}
where the KL is taken over complete outputs and equals the per-token KL summed along outputs sampled from $P^{-}_s$.
\end{proposition}
The bound follows from Pinsker's and Jensen's inequalities. It shows that forgetting is governed by the KL on the member's own prompts, which is why routing compares output distributions on stored member probes rather than on the new task alone.

\begin{proposition}[Local effect of the KL-weighted merge]
\label{prop:merge}
Assume the commit stays in the regime where each KL is well approximated by its second-order expansion. Let $\bm{F}_s$ be the Fisher information of the shared parameters on the member's prompts and $\bm{F}_t$ the Fisher information on the new task's training contexts. If $\bm{\delta}_t^{\top}\bm{F}_s\bm{\delta}_t \le c_s\,\bm{\delta}_t^{\top}\bm{F}_t\bm{\delta}_t$ and $\bar{k}_t = \tfrac{1}{2}\bm{\delta}_t^{\top}\bm{F}_t\bm{\delta}_t$, then up to third-order terms
\begin{equation}
  |\Delta_s| \le w_t \sqrt{\tfrac{c_s \bar{k}_t}{2}} \le \sqrt{\tfrac{c_s h}{8}},
  \label{eq:merge-cap}
\end{equation}
where the second inequality uses $w_t = h/(h + \bar{k}_t)$ and is tight at $\bar{k}_t = h$.
\end{proposition}
With an unweighted merge, $w_t = 1$, the bound grows as $\sqrt{c_s \bar{k}_t}$ with how far the new task drifted. The KL-weighted merge caps the local contribution of any single commit at a level set by $h$. The factor $c_s$ compares how sensitive the member's outputs are to the update with how sensitive the new task's outputs are. It is not known in advance, but it can be estimated on the stored probes after each commit. \\
\textbf{Scope:}
The analysis does not show that the training penalty bounds forgetting. The penalty and $\bar{k}_t$ are measured on the new task, whereas Proposition~\ref{prop:member-kl} requires the KL on member prompts, and the two are linked only through $c_s$. \citet{shenfeld2026rl} report that new-task KL predicts forgetting for full-parameter fine-tuning measured against the base model, and whether this carries over to intervention states measured against a shared anchor is an empirical question in this setting. Proposition~\ref{prop:merge} is local, and both results concern expected scores under sampling, so greedy decoding near a tie can change an output even when the KL is small.
\section{Proofs and scope of the theoretical analysis}
\label{sec:theory_and_proof}
\paragraph{Proof of Proposition~\ref{prop:member-kl}.}
Fix a prompt $x$. Since $G \in [0,1]$, $|\mathbb{E}_{P^{+}_s}G - \mathbb{E}_{P^{-}_s}G| \le \mathrm{TV}(P^{-}_s, P^{+}_s)$, and Pinsker's inequality gives $\mathrm{TV}(P^{-}_s, P^{+}_s) \le \sqrt{\KLdiv(P^{-}_s \,\|\, P^{+}_s)/2}$. Averaging over $x \sim \mathcal{D}_s$ and applying Jensen's inequality to the concave square root gives Eq.~\eqref{eq:member-kl}. The chain rule writes the sequence KL as $\mathbb{E}_{y \sim P^{-}_s(\cdot \mid x)} \sum_i \KLdiv\bigl(P^{-}_s(\cdot \mid x, y_{<i}) \,\|\, P^{+}_s(\cdot \mid x, y_{<i})\bigr)$, so the contexts are the member's own sampled outputs rather than teacher-forced ones. $\square$

\paragraph{Proof of Proposition~\ref{prop:merge}.}
Let $\phiparam(w) = \phiparam_{\mathcal{S}_k} + w\bm{\delta}_t$ with $\psi_s$ fixed. The member KL $\mathbb{E}_{x}\KLdiv(P_{\phiparam(0)} \,\|\, P_{\phiparam(w)})$ is zero with zero gradient at $w = 0$, and its Hessian there is the Fisher information $\bm{F}_s = \mathbb{E}_{x \sim \mathcal{D}_s}\mathbb{E}_{y \sim P^{-}_s}[\nabla \log P^{-}_s(y \mid x)\, \nabla \log P^{-}_s(y \mid x)^{\top}]$, taken with respect to the shared parameters. Hence the member KL equals $\tfrac{1}{2}w^{2}\bm{\delta}_t^{\top}\bm{F}_s\bm{\delta}_t + O(w^{3}\|\bm{\delta}_t\|^{3})$, which the domination condition bounds by $w^{2} c_s \bar{k}_t$. Proposition~\ref{prop:member-kl} then gives $|\Delta_s| \le w_t\sqrt{c_s \bar{k}_t/2}$. For $w_t = h/(h + \bar{k}_t)$, the function $\bar{k} \mapsto h\sqrt{\bar{k}}/(h + \bar{k})$ is maximized at $\bar{k} = h$ with value $\sqrt{h}/2$, which gives $\sqrt{c_s h/8}$. $\square$

\paragraph{Status of the assumptions.}
The second-order expansion holds only when $w_t\|\bm{\delta}_t\|$ is small, so Proposition~\ref{prop:merge} gives no guarantee for large updates. The factor $c_s$ is an assumption about relative sensitivity. It can be large for unrelated tasks, and routing is designed to place tasks with small output divergence on member probes together, which is consistent with a small $c_s$ but does not prove it. Proposition~\ref{prop:merge} defines $\bar{k}_t$ as the KL of the shared update alone, with the private sites held at their anchor values. A $\bar{k}_t$ computed with the trained private part is therefore a proxy for this quantity. Both propositions bound expected scores under sampling and do not apply to greedy decoding.

\paragraph{Relation to distillation and Fisher penalties.}
The penalty of Eq.~\eqref{eq:kl-control} is the output-distillation term of Learning without Forgetting~\citep{li2017learning}, with the state of the shared intervention as the teacher and new-task inputs as the contexts. To second order it equals $\tfrac{1}{2}(\phiparam - \phiparam^{\mathrm{anchor}})^{\top}\bm{F}_t(\phiparam - \phiparam^{\mathrm{anchor}})$. This is a Fisher-weighted quadratic penalty like EWC~\citep{kirkpatrick2017overcoming}, except that the Fisher is taken on the new task's contexts rather than on earlier tasks. What CRAFT adds is that the same KL also sets the merge weight, and that a related output KL on member probes drives routing.

\paragraph{The KL-regularized objective as a constrained problem.}
Any global minimizer $\phiparam_\beta$ of $\mathcal{L}_{\tau} + \beta\mathcal{L}_{\mathrm{KL}}$ also minimizes $\mathcal{L}_{\mathrm{KL}}$ among all $\phiparam$ with $\mathcal{L}_{\tau}(\phiparam) \le \mathcal{L}_{\tau}(\phiparam_\beta)$. The converse does not hold, because the set of output distributions reachable by LoReFT is not convex. In addition, the trained intervention is not a global minimizer. This view therefore describes the objective, not the trained model.

\section{Experiments}
\subsection{Benchmarks, baselines and metrics}
\label{sec:setup-exp}
TRACE~\citep{wang2023trace} chains eight tasks that differ in format and domain (~\ref{:hyper_param}), so it tests whether routing keeps dissimilar tasks apart. The 15-task stream appends seven classification tasks from \citet{wang2023orthogonal} (AG News, Amazon, DBpedia, Yahoo, Yelp, BoolQ and QQP), so it tests whether related tasks benefit from sharing as the stream grows, and the task orders of O-LoRA test sensitivity to order.
\begin{table*}[htbp]
\centering
\scriptsize
\setlength{\tabcolsep}{2.5pt}
\renewcommand{\arraystretch}{0.9}
\caption{Continual learning methods on the TRACE benchmark across foundation models. Overall Performance (OP \%, $\uparrow$), Backward Transfer (BWT \%, $\downarrow$). CRAFT averaged over two seeds and standard deviations. Best performance data in bold.}
\label{tab:trace_main}
\resizebox{\textwidth}{!}{%
\begin{tabular}{ll ccccccccc}
\toprule
 & & SeqLoRA & GEM & EWC & L2P & DualPrompt & HiDeLoRA & O-LoRA & TreeLoRA & \textbf{CRAFT (ours)} \\
\midrule
\multirow{2}{*}{\textit{Llama-2-7B-Chat}}
 & OP $\uparrow$
  & 34.30\,{\tiny$\pm$1.2} & 40.08\,{\tiny$\pm$1.6} & 42.36\,{\tiny$\pm$0.8} & 36.23\,{\tiny$\pm$0.8}
  & 37.69\,{\tiny$\pm$1.2} & 41.60\,{\tiny$\pm$0.8} & 42.78\,{\tiny$\pm$0.4} & 43.52\,{\tiny$\pm$1.0}
  & \textbf{43.69}\,{\tiny$\pm$0.39} \\
 & BWT $\downarrow$
  & 18.50\,{\tiny$\pm$0.8} & 6.77\,{\tiny$\pm$1.2} & 5.97\,{\tiny$\pm$0.8} & 8.25\,{\tiny$\pm$0.8}
  & 8.03\,{\tiny$\pm$0.8} & 7.12\,{\tiny$\pm$0.4} & 7.16\,{\tiny$\pm$0.4} & 3.46\,{\tiny$\pm$0.4}
  & \textbf{2.29}\,{\tiny$\pm$0.3} \\
\midrule
\multirow{2}{*}{\textit{Gemma-2B-it}}
 & OP $\uparrow$
  & 31.89\,{\tiny$\pm$0.8} & 26.48\,{\tiny$\pm$1.5} & 28.35\,{\tiny$\pm$1.6} & 31.14\,{\tiny$\pm$1.2}
  & 32.42\,{\tiny$\pm$1.0} & 33.25\,{\tiny$\pm$0.9} & 33.73\,{\tiny$\pm$0.8} & 33.41\,{\tiny$\pm$0.9}
  & \textbf{35.18}\,{\tiny$\pm$0.44} \\
 & BWT $\downarrow$
  & 15.28\,{\tiny$\pm$0.4} & 18.25\,{\tiny$\pm$0.9} & 16.96\,{\tiny$\pm$1.2} & 15.77\,{\tiny$\pm$0.7}
  & 14.25\,{\tiny$\pm$0.5} & 13.66\,{\tiny$\pm$0.5} & 12.36\,{\tiny$\pm$0.4} & 8.50\,{\tiny$\pm$0.5}
  & \textbf{2.47}\,{\tiny$\pm$0.17} \\
\midrule
\multirow{2}{*}{\textit{Llama-3.2-1B-Instruct}}
 & OP $\uparrow$
  & 29.73\,{\tiny$\pm$1.6} & 32.19\,{\tiny$\pm$2.0} & 31.96\,{\tiny$\pm$1.6} & 29.38\,{\tiny$\pm$1.2}
  & 30.76\,{\tiny$\pm$1.2} & 33.73\,{\tiny$\pm$1.2} & 32.94\,{\tiny$\pm$0.8} & 36.14\,{\tiny$\pm$0.7}
   & \textbf{39.45}\,{\tiny$\pm$1.05} \\
 & BWT $\downarrow$
  & 17.03\,{\tiny$\pm$1.2} & 10.74\,{\tiny$\pm$1.6} & 11.62\,{\tiny$\pm$1.2} & 13.57\,{\tiny$\pm$0.8}
  & 11.34\,{\tiny$\pm$0.8} & 12.36\,{\tiny$\pm$0.8} & 12.89\,{\tiny$\pm$1.2} & 7.36\,{\tiny$\pm$0.8}
   & \textbf{2.38}\,{\tiny$\pm$0.26} \\
\bottomrule
\end{tabular}%
}
\end{table*}
\begin{table*}[htbp]
\centering
\scriptsize
\setlength{\tabcolsep}{3pt}
\caption{Results on the 15-task benchmark, the eight TRACE tasks followed by seven classification tasks (DBpedia, Amazon, Yahoo, AG News, Yelp, BoolQA and QQP), with Llama-3.2-1B-Instruct. OP $\uparrow$ and BWT $\downarrow$ in \%. Best result per row in bold.}
\label{tab:long15}
\begin{tabular}{l cccccccccc}
\toprule
 & SeqLoRA & OGD & GEM & EWC & L2P & DualPrompt & HiDeLoRA & O-LoRA & TreeLoRA & \textbf{CRAFT (ours)} \\
\midrule
OP $\uparrow$    & 40.71 & 32.52 & 35.48 & 31.46 & 41.05 & 41.29 & 42.38 & 44.02 & 45.68 & \textbf{47.56} \\
BWT $\downarrow$ & 15.72 & 21.32 & 18.33 & 22.22 & 14.92 & 15.58 & 11.23 & 10.99 & 6.41  & \textbf{4.30} \\
\bottomrule
\end{tabular}
\end{table*}
\paragraph{Protocol.} Results use the test split, except the ablations and the sensitivity study, which use the validation split of TRACE-8 with Llama-3.2-1B-Instruct. TRACE uses 5{,}000 training examples per task and the 15-task streams 500. CRAFT results in Table~\ref{tab:trace_main} are the mean and standard deviation over two seeds; the task orders and the Math-LLM stream use two seeds, Tables~\ref{tab:components}(a) and~\ref{tab:private_ablation} average two seeds, and all other analyses use one seed. \\
\textbf{Baselines:} CRAFT is compared with SeqLoRA, GEM~\citep{lopez2017gradient}, EWC~\citep{kirkpatrick2017overcoming}, L2P~\citep{wang2022learning}, DualPrompt~\citep{wang2022bDualPrompt}, HiDeLoRA~\citep{wang2025hide}, O-LoRA~\citep{wang2023orthogonal} and TreeLoRA~\citep{qian2025tree}, and OGD~\citep{farajtabar2020ogd} on the 15-task stream (~\ref{:baselines}). None of them uses task identity at test time. CRAFT results use task-agnostic routing, which needs no task identity (Table~\ref{tab:routing_ablation}). \\
\textbf{Metrics:} With $a_{j,t}$ the score on task $t$ after training up to task $j$, $\mathrm{OP} = \frac{1}{T}\sum_{t=1}^{T} a_{T,t}$ and $\mathrm{BWT} = \frac{1}{T}\sum_{t=1}^{T}\max(a_{t,t} - a_{T,t}, 0)$, where, following the TreeLoRA implementation, the term of a task that improves is set to zero. A positive BWT means forgetting, and higher OP and lower BWT are better.
\subsection{Comparison with CL baselines}
\label{sec:main-results}
CRAFT has the lowest BWT and the highest OP on all three backbones of Table~\ref{tab:trace_main}. On Llama-3.2-1B-Instruct, Gemma-2B-it and Llama-2-7B-Chat, it forgets 4.98, 6.03 and 1.17 points less than TreeLoRA, the baseline that forgets least, and leads the strongest baseline in OP by 3.31, 1.45 and 0.17 points, the last within one standard deviation. The gain comes from retention: every baseline updates parameters that all earlier tasks depend on, while CRAFT changes only the anchored shared intervention of the arriving task. The lead is therefore small on Llama-2-7B-Chat, where TreeLoRA forgets only 3.46 points, and on Gemma-2B-it the large cut in forgetting yields a smaller OP gain, which suggests that learning each task, rather than retaining it, limits CRAFT there. \\
\textbf{Longer task sequence:}
On the 15-task stream, CRAFT leads TreeLoRA by 1.88 OP points with 2.11 points less forgetting (Table~\ref{tab:long15}). Its BWT of 4.30 is higher than on TRACE because 11 of the 15 tasks join an existing shared intervention, and only a join can cause forgetting (Section~\ref{sec:exp-routing}).
\\ \textbf{Six task orders:}
The six task orders of O-LoRA~\citep{wang2023orthogonal} arrange four tasks in orders 1 to 3 (DBpedia, Amazon, Yahoo and AG News) and fifteen in orders 4 to 6 (MNLI, CB, WiC, COPA, QQP, BoolQA, RTE, IMDB, Yelp, Amazon, SST-2, DBpedia, AG News, MultiRC and Yahoo), with baselines from \citet{qian2025tree}. CRAFT has the highest OP and the lowest BWT in every order, leading the strongest baseline by 4.9 to 20.9 OP points on orders 1 to 3 ( Table~\ref{tab:olora_orders_full}) and by 8.2 to 10.9 points, with 0.4 to 1.9 points less forgetting, on orders 4 to 6 (Table~\ref{tab:olora_orders_long}). \\
\textbf{Math reasoning:}
On the Math-LLM stream, each task opens its own shared intervention, so no task can forget, and CRAFT leads TreeLoRA by 10.5 OP points with zero BWT ( Table~\ref{tab:math_llm_comparison_treelora}).
\begin{figure}[t] \centering \includegraphics[width=0.5\textwidth]{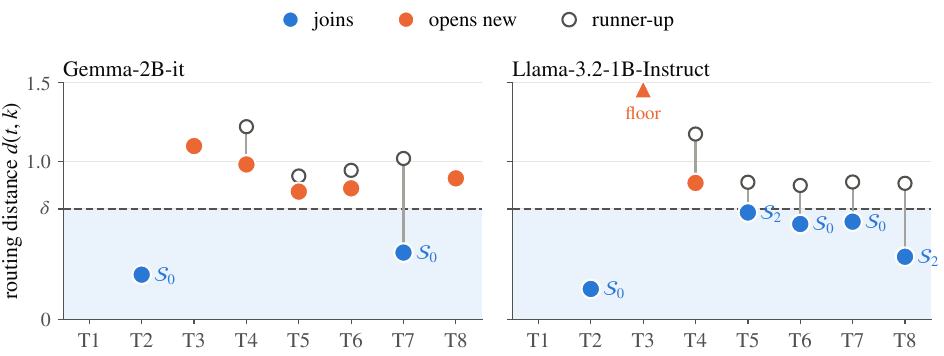} \caption{Routing distance of each TRACE-8 task to every existing shared intervention, in arrival order. A task joins its closest shared intervention below $\delta = 0.7$ (shaded) and otherwise opens a new one.} \label{fig:routing_distance} \end{figure} Figure~\ref{fig:routing_distance} shows every routing decision. On Gemma-2B-it, five tasks lie above $\delta$ from every shared intervention and open their own, while on Llama-3.2-1B-Instruct five tasks join a shared intervention.

\begin{figure}[t]
\centering
\includegraphics[width=0.8\columnwidth]{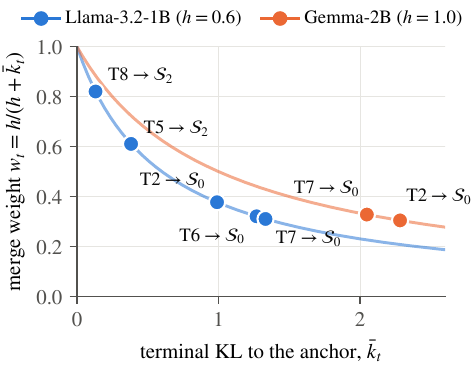}
\caption{Terminal KL of each joining task to its anchor and the merge weight it receives on TRACE-8. Curves show $w_t = h/(h+\bar{k}_t)$, and each point is a task joining the shared intervention after the arrow. Tasks follow the numbering of Figure~\ref{fig:routing_distance}.}
\label{fig:merge_pull}
\end{figure}

\subsection{Routing confines forgetting to shared interventions}
\label{sec:exp-routing}
As Section~\ref{sec:theory} predicts, a task cannot be changed by training in another shared intervention. Forgetting therefore arises only when a later task joins the same shared intervention, the event that the KL anchor and the KL-weighted merge control. \\
\textbf{Routing follows output agreement}\\
On TRACE-8 with Llama-3.2-1B-Instruct, routing forms three shared interventions: C-STANCE, FOMC and both NumGLUE tasks, which all answer with a short label or number; MeetingBank alone; and Py150, ScienceQA and 20Minuten. Joins are mostly clear, as 20Minuten joins at a distance of 0.40 while the next candidate lies at 0.86. MeetingBank opens its own shared intervention through the no-signal floor, because on its prompts the warmed-up task and the closest shared state give almost identical outputs, so the distance carries no evidence. \\ 
\textbf{Sharing transfers knowledge:}
A later task can also improve an earlier one. On the 15-task stream ( Table~\ref{tab:eval_matrix_15task}), C-STANCE rises from 51.0 to 56.0 when FOMC joins its shared intervention, and Yelp rises from 42.0 to 46.0 when QQP joins and keeps the gain to the end. Per-task storage cannot produce such gains, and a single intervention for all tasks loses them to interference (Section~\ref{sec:exp-loreft}). \\ 
\textbf{Task identification at test time:}
Table~\ref{tab:routing_ablation} compares ways of selecting the stored state at test time. Without the task ID, CRAFT identifies every test task and loads the shared intervention that holds it, so it matches the task-aware lookup, the upper bound for any selection over the same states, at 39.45 OP and 2.38 BWT. Uniform averaging mixes edits learned for different output formats and reaches only 32.4 OP, and perplexity ranking picks the same shared state for every task, so it selects correctly only for the three tasks stored there and reaches 25.6 to 31.3 OP. Even with the task ID, the shared state alone gives 33.8 OP, so the private l-stream carries 5.7 points, and the task itself, not only its shared intervention, must be identified.
\subsection{Representation fine-tuning as a continual learner}
\label{sec:exp-loreft}
A frozen backbone still allows interference inside an intervention that several tasks share. A single intervention trained on all TRACE-8 tasks forgets most of what it learns, with BWT of 33.1 on Llama-3.2-1B-Instruct and 26.9 on R1-Distill-Qwen-1.5B, because tasks with different output formats pull one subspace in incompatible directions (Table~\ref{tab:unit_of_sharing}). A KL anchor lowers this to 12.8 and 14.7 points, but one state still serves every format. CRAFT addresses both causes, as routing keeps incompatible formats apart and the anchor protects each shared intervention, so BWT falls to 2.4 and 0.8 points and OP rises by 14.3 and 12.7 points over the anchored single intervention.
\begin{table}[t]
\centering
\footnotesize
\setlength{\tabcolsep}{4pt}
\caption{LoReFT trained on TRACE-8, OP / BWT. All-in-one: one intervention for all tasks. No reg.: no regularizer. CRAFT: routed shared interventions with the KL anchor.}
\label{tab:unit_of_sharing}
\resizebox{\columnwidth}{!}{%
\begin{tabular}{l cc c}
\toprule
 & \multicolumn{2}{c}{All-in-one} & \\
\cmidrule(lr){2-3}
Backbone & No reg. & KL & \textbf{CRAFT} \\
\midrule
Llama-3.2-1B  & 15.09 / 33.08 & 25.11 / 12.83 & \textbf{39.45 / 2.38} \\
R1-Qwen-1.5B  & 18.77 / 26.85 & 29.10 / 14.71 & \textbf{41.78 / 0.77} \\
\bottomrule
\end{tabular}}
\end{table}
\begin{table}[t]
\centering
\footnotesize
\setlength{\tabcolsep}{3pt}
\caption{Selecting the stored state at test time on TRACE-8 with Llama-3.2-1B-Instruct. Task ident.: share of tasks (\%) for which the selected shared intervention is the one that holds the task. Shared state only, uniform and top-$k$ omit the private part. Top-$k$ keeps the $k$ shared states with the lowest perplexity on 16 unlabeled test prompts.}
\label{tab:routing_ablation}
\resizebox{\columnwidth}{!}{%
\begin{tabular}{l c c c c}
\toprule
Selection & ID given & T Ident.\ $\uparrow$ & OP $\uparrow$ & BWT $\downarrow$ \\
\midrule
Oracle, Task-aware lookup                & yes & --   & 39.45 & 2.38 \\
Task-aware, shared state    & yes & --   & 33.77 & 7.12 \\
\midrule
Uniform over shared states       & no  & --   & 32.44 & 12.38 \\
Top-$k$ by perplexity, $k=1$     & no  & 37.5 & 25.55 & 15.49 \\
Top-$k$ by perplexity, $k=2$     & no  & 37.5 & 30.26 & 12.97 \\
Top-$k$ by perplexity, $k=3$     & no  & 37.5 & 31.33 & 12.72 \\
\textbf{CRAFT}           & no  & \textbf{100} & \textbf{39.45} & \textbf{2.38} \\
\bottomrule
\end{tabular}}
\end{table}
\begin{table}[htbp]
\centering
\caption{Results on task orders 4 to 6 of O-LoRA~\citep{wang2023orthogonal}, which arrange fifteen classification tasks, with Llama-3.2-1B-Instruct as the foundation model. Each entry reports OP $\uparrow$ / BWT $\downarrow$ (\%). Best result per order in bold.}
\label{tab:olora_orders_long}
\small
\setlength{\tabcolsep}{3pt}
\resizebox{\columnwidth}{!}{%
\begin{tabular}{l c c c c c}
\toprule
Order & FIX & HiDeLoRA & O-LoRA & TreeLoRA & \textbf{CRAFT (ours)} \\
\midrule
Order 4 & 52.13 / -- & 59.44 / 4.33 & 59.89 / 4.67 & 58.45 / 4.98 & \textbf{68.13}\,{\tiny$\pm$2.67} / \textbf{2.79} \\
Order 5 & 52.13 / -- & 54.49 / 7.52 & 57.05 / 4.42 & 58.12 / 3.31 & \textbf{66.57}\,{\tiny$\pm$2.70} / \textbf{2.95} \\
Order 6 & 52.13 / -- & 57.26 / 6.98 & 58.02 / 4.73 & 59.00 / 4.12 & \textbf{69.92}\,{\tiny$\pm$0.40} / \textbf{3.15} \\
\bottomrule
\end{tabular}}
\end{table}
\begin{table*}[htbp]
\centering
\footnotesize
\caption{Performance comparison on TRACE with Llama-2-7B-Chat. For a fair comparison, generative replay is not employed in SAPT.  CRAFT as in Table~\ref{tab:trace_main}.}
\label{tab:sapt_tasl_comparison}
\setlength{\tabcolsep}{3pt}
\resizebox{\textwidth}{!}{%
\begin{tabular}{lcccccccccc}
\toprule
 & \textbf{GEM} & \textbf{EWC} & \textbf{L2P}
 & \textbf{DualPrompt} & \textbf{HiDeLoRA} & \textbf{O-LoRA}
 & \textbf{TreeLoRA} & \textbf{SAPT} & \textbf{TASL} & \textbf{CRAFT (ours)} \\
\midrule
OP (\%) $\uparrow$    & 40.08 & 42.36 & 36.23 & 37.69 & 41.60 & 42.78 & 43.52 & 42.93 & 43.19 & \textbf{43.69} \\
BWT (\%) $\downarrow$ &  \phantom{0}6.77 & \phantom{0}5.97 & \phantom{0}8.25 & \phantom{0}8.03 & \phantom{0}7.12 & \phantom{0}7.16 & \phantom{0}3.46 & \phantom{0}5.49 & \phantom{0}4.58 & \textbf{2.29} \\
\bottomrule
\end{tabular}}
\end{table*}

\subsection{KL anchor and KL-weighted merge}
\label{sec:kl-dynamics}
Table~\ref{tab:components} measures forgetting inside shared interventions by $\mathrm{BWT}_{\mathrm{in}}$, the BWT over tasks that share their intervention with a later task. Without the anchor, a joining task overwrites the shared state and $\mathrm{BWT}_{\mathrm{in}}$ rises from 1.8 to 21.6 points. The anchor alone is not enough: with an overwrite instead of the KL-weighted merge, $\mathrm{BWT}_{\mathrm{in}}$ stays at 13.7, because a task that drifted far is still committed in full. With both components, OP is 5.1 points higher than without the anchor.
The merge half-life $h$ shows the same mechanism as a dose response: as $h$ grows from 0.2 to 5, the merge commits more of a task that drifted far, and $\mathrm{BWT}_{\mathrm{in}}$ rises from 0.5 to 9.4 points, then to 13.1 for an overwrite. OP peaks at $h = 0.6$, because a smaller $h$ also commits too little of the new task (~\ref{sec:kl-forgetting-empirical} relates terminal KL to forgetting).
\begin{table}[t]
\centering
\footnotesize
\setlength{\tabcolsep}{3.5pt}
\caption{KL anchor and KL-weighted merge, Llama-3.2-1B-Instruct, TRACE-8 validation split, in the ablation configuration of ~\ref{sec:app-ablation}, which keeps the same partition in every row. (a) Mean of seeds 42 and 1234. (b) Merge half-life $h$, seed 42. $\mathrm{BWT}_{\mathrm{in}}$: BWT over tasks that share their intervention with a later task.}
\label{tab:components}
\begin{tabular}{l c c c}
\toprule
 & OP $\uparrow$ & BWT $\downarrow$ & $\mathrm{BWT}_{\mathrm{in}}$ $\downarrow$ \\
\midrule
\multicolumn{4}{l}{\textit{(a) Components, two seeds}} \\
\textbf{CRAFT}                         & \textbf{46.70} & \textbf{0.46} & \textbf{1.85} \\
No KL anchor                           & 41.57 & 5.39 & 21.55 \\
KL anchor, overwrite merge             & 43.24 & 3.44 & 13.74 \\
\midrule
\multicolumn{4}{l}{\textit{(b) Merge half-life $h$, seed 42}} \\
$h = 0.2$                              & 45.61 & 0.14 & 0.54 \\
$h = 0.6$ (CRAFT)                      & 45.94 & 0.40 & 1.60 \\
$h = 2.0$                              & 44.54 & 1.43 & 5.72 \\
$h = 5.0$                              & 44.67 & 2.34 & 9.38 \\
Overwrite                              & 42.56 & 3.27 & 13.09 \\
\bottomrule
\end{tabular}
\end{table}
\subsection{Robustness and scalability}
\label{sec:robustness}
\textbf{Stream length and storage:}
As the stream grows from 8 to 15 tasks, the number of shared interventions grows only from three to four, and the fifteen-task orders form six to eight (~\ref{sec:routing-robustness-app}). The stored state therefore grows mainly with the small private part of each task: on the 15-task stream, CRAFT stores 6.36M parameters on Llama-3.2-1B-Instruct against 15.73M for one intervention per task and 12.78M for one LoRA adapter on $q,v$ per task \\
\textbf{Model scale and cost:}
The routing configuration transfers from 1B to 7B parameters with only the learning rate, and on Gemma-2B-it the anchor, set per backbone. On Llama-3.2-1B-Instruct, CRAFT trains TRACE-8 in 5.9 A100 GPU-hours against 5.2 for one intervention per task.
\subsection{Recent baselines}
CRAFT is compared with two recent continual learning methods, SAPT~\citep{zhao2024sapt} and TASL~\citep{feng2024tasl}, on Llama-2-7B-Chat using the TRACE benchmark in Table~\ref{tab:sapt_tasl_comparison}. SAPT is used in its replay-free variant. TASL localizes a small subset of parameters carrying each task's skill and consolidates them across the sequence, restricting updates to task-relevant subsets to reduce overwriting.

\subsection{Task orders 1 to 3}
\begin{table*}[t]
\centering
\caption{Results on task orders 1 to 3 of O-LoRA~\citep{wang2023orthogonal}, which arrange four classification tasks (DBpedia, Amazon, Yahoo and AG News), with Llama-3.2-1B-Instruct as the foundation model. Each entry reports OP $\uparrow$ / BWT $\downarrow$ (\%). Best result per order in bold.}
\label{tab:olora_orders_full}
\small
\setlength{\tabcolsep}{3pt}
\begin{tabular}{l c c c c c c c c}
\toprule
Order & FIX & OGD & GEM & SeqLoRA & HiDeLoRA & O-LoRA & TreeLoRA & \textbf{CRAFT (ours)} \\
\midrule
Order 1 & 48.75 / -- & 54.16 / 10.82 & 54.10 / 10.25 & 55.71 / 8.63 & 56.32 / 2.75 & 59.50 / 2.51 & 59.73 / 2.22 & \textbf{76.13}\,{\tiny$\pm$0.53} / \textbf{0.63} \\
Order 2 & 48.75 / -- & 46.82 / 21.50 & 46.70 / 20.93 & 45.26 / 7.70 & 53.41 / 5.58 & 52.53 / 5.65 & 53.78 / 5.74 & \textbf{74.63}\,{\tiny$\pm$3.01} / \textbf{0.13} \\
Order 3 & 48.75 / -- & 43.79 / 27.31 & 51.12 / 16.24 & 49.03 / 19.05 & 61.25 / 3.12 & 63.82 / 2.03 & 62.76 / 2.23 & \textbf{68.75}\,{\tiny$\pm$0.00} / \textbf{0.88} \\
\bottomrule
\end{tabular}
\end{table*}

\subsection{Math reasoning}
The Math-LLM benchmark is a curated subset of three math and reasoning tasks: ScienceQA, NumGLUE-cm, and NumGLUE-ds. Table~\ref{tab:math_llm_comparison_treelora} reports results with Llama-2-7B-Chat. Routing gives each of the three tasks its own shared intervention, so no task can forget, and CRAFT reaches the highest OP at 56.13, 10.5 points above TreeLoRA, with zero BWT on both seeds.
\begin{table*}[htbp]
\centering
\caption{Continual learning comparison on the Math-LLM benchmark (ScienceQA, NumGLUE-cm, NumGLUE-ds) using Llama-2-7B-Chat as the foundation model. Overall performance (OP, \%, $\uparrow$) and backward transfer (BWT, \%, $\downarrow$). CRAFT: mean over two seeds and 4, 5 and 5 epochs, learning rate $10^{-4}$. Best per row in bold.}
\label{tab:math_llm_comparison_treelora}
\footnotesize
\setlength{\tabcolsep}{3pt}
\resizebox{\textwidth}{!}{%
\begin{tabular}{l c c c c c c c c c}
\toprule
 & SeqLoRA & GEM & EWC & L2P & DualPrompt & HiDeLoRA & O-LoRA & TreeLoRA & CRAFT (ours) \\
\midrule
OP (\%) $\uparrow$  & 40.24$\pm$1.0 & 40.75$\pm$1.4 & 37.25$\pm$1.1 & 31.67$\pm$1.0 & 37.72$\pm$1.0 & 42.83$\pm$0.7 & 42.00$\pm$1.0 & 45.59$\pm$1.2 & \textbf{56.13$\pm$1.43} \\
BWT (\%) $\downarrow$ & 9.12$\pm$0.8 & 16.25$\pm$0.5 & 17.50$\pm$0.6 & 2.54$\pm$0.4 & 9.54$\pm$0.7 & 11.92$\pm$0.5 & 11.67$\pm$0.6 & 2.32$\pm$0.6 & \textbf{0.00$\pm$0.00} \\
\bottomrule
\end{tabular}}%
\end{table*}

\subsection{Per-task evaluation matrix}
\label{app:eval-matrix}
Table~\ref{tab:eval_matrix_15task} gives the score on every earlier task after each training step of the 15-task stream. A column changes only in the rows where a later task joins the shared intervention of that column's task. Every other entry repeats exactly from row to row, which is the isolation across shared interventions of Section~\ref{sec:theory}.

% Fill from <run>/eval_test_bwd/full_grid.csv of the FULLGRID job for the 15-task run.

\section{Routing analysis}
\subsection{Routing decisions}
Tables~\ref{tab:routing_diag_trace8} and~\ref{tab:routing_diag_15task} list every routing decision of CRAFT on TRACE-8 with Gemma-2B-it and on order 6 of the O-LoRA 15-task pool with Llama-3.2-1B-Instruct, seed 42, with no limit on the size of a shared intervention. For each task, the tables give the closest shared intervention, its distance $d(t,k^\star)$ from Eq.~\eqref{eq:route_dist}, and the runner-up. On TRACE-8, only FOMC and NumGLUE-ds join a shared intervention, both the one of C-STANCE, and NumGLUE-ds joins at 0.42 against 1.02 for the runner-up. The other five tasks lie 0.81 to 1.10 from their closest shared intervention, above $\delta = 0.7$, so each opens its own. On order 6, twelve of the fifteen tasks join a shared intervention. Of the eleven joins with a runner-up, ten have it at least 0.2 farther away, and Yahoo is the one close call, at 0.318 against 0.320. A wrong split costs one extra shared state, while a wrong join can cause forgetting inside a shared intervention.
\begin{table}[h]
\centering
\small
\setlength{\tabcolsep}{4pt}
\caption{Routing decisions on TRACE-8, Gemma-2B-it, $\delta = 0.7$, seed 42. Final shared interventions: $\mathcal{S}_0$ = \{C-STANCE, FOMC, NumGLUE-ds\}, $\mathcal{S}_1$ = \{MeetingBank\}, $\mathcal{S}_2$ = \{Py150\}, $\mathcal{S}_3$ = \{ScienceQA\}, $\mathcal{S}_4$ = \{NumGLUE-cm\}, $\mathcal{S}_5$ = \{20Minuten\}.}
\label{tab:routing_diag_trace8}
\resizebox{\columnwidth}{!}{%
\begin{tabular}{l l c c c c}
\toprule
Task & Decision & $k^\star$ & $d(t,k^\star)$ & Runner-up & $d$ \\
\midrule
C-STANCE    & new $\mathcal{S}_0$  & -- & --    & -- & --    \\
FOMC        & join $\mathcal{S}_0$ & 0  & 0.283 & -- & --    \\
MeetingBank & new $\mathcal{S}_1$  & 0  & 1.099 & -- & --    \\
Py150       & new $\mathcal{S}_2$  & 1  & 0.982 & 0  & 1.221 \\
ScienceQA   & new $\mathcal{S}_3$  & 0  & 0.809 & 1  & 0.909 \\
NumGLUE-cm  & new $\mathcal{S}_4$  & 0  & 0.830 & 1  & 0.944 \\
NumGLUE-ds  & join $\mathcal{S}_0$ & 0  & 0.423 & 1  & 1.020 \\
20Minuten   & new $\mathcal{S}_5$  & 1  & 0.894 & 0  & 0.904 \\
\bottomrule
\end{tabular}}
\end{table}

\begin{table}[h]
\centering
\small
\setlength{\tabcolsep}{4pt}
\caption{Routing decisions on order 6 of the O-LoRA 15-task stream, Llama-3.2-1B-Instruct, $\delta = 0.7$, no capacity limit, seed 42. Final shared interventions: $\mathcal{S}_0$ = \{Yelp, Amazon, Yahoo\}, $\mathcal{S}_1$ = \{MNLI, CB, RTE\}, $\mathcal{S}_2$ = \{COPA, QQP, IMDB, SST-2, DBpedia, AG News, MultiRC, BoolQA, WiC\}.}
\label{tab:routing_diag_15task}
\resizebox{\columnwidth}{!}{%
\begin{tabular}{l l c c c c}
\toprule
Task & Decision & $k^\star$ & $d(t,k^\star)$ & Runner-up & $d$ \\
\midrule
Yelp     & new $\mathcal{S}_0$  & -- & --    & -- & --    \\
Amazon   & join $\mathcal{S}_0$ & 0  & 0.473 & -- & --    \\
MNLI     & new $\mathcal{S}_1$  & 0  & 0.876 & -- & --    \\
CB       & join $\mathcal{S}_1$ & 1  & 0.200 & 0  & 1.144 \\
COPA     & new $\mathcal{S}_2$  & 0  & 0.768 & 1  & 7.781 \\
QQP      & join $\mathcal{S}_2$ & 2  & 0.505 & 0  & 0.782 \\
RTE      & join $\mathcal{S}_1$ & 1  & 0.247 & 2  & 0.571 \\
IMDB     & join $\mathcal{S}_2$ & 2  & 0.586 & 0  & 1.014 \\
SST-2    & join $\mathcal{S}_2$ & 2  & 0.237 & 0  & 0.759 \\
DBpedia  & join $\mathcal{S}_2$ & 2  & 0.386 & 1  & 0.774 \\
AG News  & join $\mathcal{S}_2$ & 2  & 0.246 & 0  & 0.447 \\
Yahoo    & join $\mathcal{S}_0$ & 0  & 0.318 & 2  & 0.320 \\
MultiRC  & join $\mathcal{S}_2$ & 2  & 0.288 & 0  & 0.631 \\
BoolQA   & join $\mathcal{S}_2$ & 2  & 0.214 & 0  & 0.496 \\
WiC      & join $\mathcal{S}_2$ & 2  & 0.696 & 0  & 0.941 \\
\bottomrule
\end{tabular}}
\end{table}
\subsection{Routing across task orders}
\label{sec:routing-robustness-app}
Table~\ref{tab:routing_orders} reports the shared interventions that routing forms on the three 15-task orders of O-LoRA with Llama-3.2-1B-Instruct. The number of shared interventions stays within a narrow range across orders, so the partition depends on the tasks rather than on the order in which they arrive.
\begin{table}[ht]
\centering
\footnotesize
\setlength{\tabcolsep}{3pt}
\caption{Shared interventions formed on orders 4 to 6 of the O-LoRA 15-task pool, Llama-3.2-1B-Instruct. $K$: number of shared interventions for seeds 42 / 1234; largest: tasks in the largest one. OP and BWT: mean over the two seeds, as in Table~\ref{tab:olora_orders_long}.}
\label{tab:routing_orders}
\begin{tabular}{l c c c c}
\toprule
Order & $K$ & Largest & OP $\uparrow$ & BWT $\downarrow$ \\
\midrule
4 & 8 / 6 & 4 & 68.13 & 2.79 \\
5 & 8 / 8 & 4 & 66.57 & 2.95 \\
6 & 6 / 6 & 4 & 69.92 & 3.15 \\
\bottomrule
\end{tabular}
\end{table}

\subsection{Routing threshold $\delta$}
\label{sec:delta-app}
\begin{table}[ht]
\centering
\small
\caption{Routing threshold $\delta$, TRACE-8 validation split, Llama-3.2-1B-Instruct, seed 42, ablation configuration. Final value marked $\star$.}
\label{tab:delta_nocap}
\begin{tabular}{c c c c c}
\toprule
$\delta$ & $K$ & LA $\uparrow$ & OP $\uparrow$ & BWT $\downarrow$ \\
\midrule
0.3 & 7 & 45.50 & 45.44 & 0.06 \\
0.5 & 7 & 44.22 & 44.21 & 0.01 \\
0.7 & 6 & \textbf{46.34} & \textbf{45.94} & 0.40 \\
1.0 & 5 & 42.56 & 41.95 & 0.62 \\
1.5 & 5 & 42.85 & 42.13 & 0.72 \\
2.0 & 5 & 42.48 & 41.89 & 0.59 \\
\bottomrule
\end{tabular}
\end{table}

\paragraph{Separation of distant tasks.}
FOMC and ScienceQA have very different output distributions. On Llama-3.2-1B-Instruct, ScienceQA lies 1.44 from the shared intervention of FOMC when the two tasks form a stream on their own, and 1.34 to 1.43 from the shared intervention that holds C-STANCE and FOMC in the TRACE-8 runs of Table~\ref{tab:delta_nocap}. ScienceQA therefore stays apart from FOMC at the default $\delta = 0.7$ and at $\delta = 1.0$, and could join it only from $\delta = 1.5$, where routing still places it in a closer shared intervention.

\subsection{Routing direction}
\label{sec:routing-direction}
Eq.~\eqref{eq:route_dist} measures the distance on the probes of the shared intervention's members. A variant measures the same three divergences on the arriving task's probe $\mathcal{P}_{t}$ only,
\begin{equation*}
  d_{\mathrm{fwd}}(t,k) = \frac{D_{KS_k}}{\max\bigl(\min(D_{K}, D_{S_k}),\, \varepsilon_{\mathrm{sig}}\bigr)}.
\end{equation*}
The arriving-task variant asks whether the shared intervention already behaves like the new task on the new task's data. The member-side distance asks whether the new task would change what the shared intervention does on its own data, which is the quantity that forgetting depends on. On the forced-join pairs of TRACE-8 tasks, each scored by the drop that the second task causes on the first when the two share an intervention, the member-side distance ranks this damage with Spearman $\rho = 0.50$ on R1-Distill-Qwen-1.5B ($p = 0.007$) against $\rho = -0.24$ for the arriving-task variant; on Llama-3.2-1B-Instruct neither direction ranks it significantly ($\rho = 0.14$ and $-0.13$). Inside the full method, member-side routing shares one more pair of tasks, learns each task better and reaches 4.1 points higher OP, at a BWT of 0.40 against 0.05 (Table~\ref{tab:route_dir}).
\begin{table}[h]
\centering
\small
\caption{Arriving-task and member-side routing distances on TRACE-8 with Llama-3.2-1B-Instruct, in the ablation configuration.}
\label{tab:route_dir}
\begin{tabular}{l c c c c c}
\toprule
Direction & $\delta$ & $K$ & LA $\uparrow$ & OP $\uparrow$ & BWT $\downarrow$ \\
\midrule
Arriving task & 0.7 & 7 & 41.93 & 41.88 & \textbf{0.05} \\
Member-side   & 0.7 & 6 & \textbf{46.34} & \textbf{45.94} & 0.40 \\
\bottomrule
\end{tabular}
\end{table}

\section{Ablations and sensitivity}
\label{sec:app-ablation}
\subsection{Hyperparameter sensitivity}
\begin{table}[t]
\centering
\small
\setlength{\tabcolsep}{4pt}
\caption{Hyperparameter sensitivity, Llama-3.2-1B-Instruct, TRACE-8 validation split, seed 42, ablation configuration. One value changes at a time; final values marked $\star$. $K$: number of shared interventions.}
\label{tab:sensitivity}
\resizebox{\columnwidth}{!}{%
\begin{tabular}{l l c c c c}
\toprule
Hyperparameter & Value & $K$ & LA $\uparrow$ & OP $\uparrow$ & BWT $\downarrow$ \\
\midrule
Anchor strength $\beta$ & none & 6 & 46.08 & 40.75 & 5.32 \\
 & 0.1 & 6 & 46.44 & 46.26 & 0.24 \\
 & 0.3 $\star$ & 6 & 46.34 & 45.94 & 0.40 \\
 & 1.0 & 7 & 48.12 & 47.93 & 0.19 \\
\midrule
Merge half-life $h$ & 0.2 & 6 & 45.75 & 45.61 & 0.14 \\
 & 0.6 $\star$ & 6 & 46.34 & 45.94 & 0.40 \\
 & 2.0 & 6 & 45.97 & 44.54 & 1.43 \\
 & 5.0 & 6 & 47.01 & 44.67 & 2.34 \\
 & overwrite & 6 & 45.83 & 42.56 & 3.27 \\
\midrule
Intervention rank $r$ & 4 & 6 & 45.79 & 45.25 & 0.54 \\
 & 8 $\star$ & 6 & 46.34 & 45.94 & 0.40 \\
 & 16 & 6 & 41.14 & 41.16 & 0.07 \\
\midrule
Positions per stream $t_{pos}$ & 9 & 7 & 45.94 & 45.82 & 0.12 \\
 & 15 $\star$ & 6 & 46.34 & 45.94 & 0.40 \\
 & 21 & 6 & 47.21 & 46.92 & 0.32 \\
\midrule
Warm-up steps $S_{\mathrm{wu}}$ & 50 & 6 & 46.83 & 46.30 & 0.53 \\
 & 100 $\star$ & 6 & 46.34 & 45.94 & 0.40 \\
 & 400 & 4 & 45.28 & 44.08 & 1.20 \\
\bottomrule
\end{tabular}}
\end{table}

\textbf{Anchor strength $\beta$:} From $\beta = 0.1$ to $1.0$, BWT stays between 0.19 and 0.40 and OP between 45.9 and 47.9. Removing the anchor raises BWT to 5.32 and lowers OP by more than five points. A weak anchor already does the work, because its role is to keep the joining task near the state of its shared intervention, not to freeze it.

\textbf{Merge half-life $h$:} BWT rises steadily as $h$ grows, from 0.14 at $h = 0.2$ to 2.34 at $h = 5$ and 3.27 for an overwrite, because a larger $h$ commits more of the joining task's update into the members' shared state. OP is flat from 0.2 to 0.6, and $h = 0.6$ gives the highest OP, keeping enough of the joining task's learning while protecting the members.

\textbf{Rank, positions and warm-up:} Ranks 4 and 8 and 9 to 21 positions per stream give the same OP within one point. Rank 16 learns each task less well at the shared learning rate, with LA 41.1 against 46.3, while it forgets almost nothing, so its lower OP comes from learning rather than from forgetting. A 400-step warm-up moves each warmed-up task further from the unadapted model, which shrinks the routing distance and joins tasks that a 100-step warm-up keeps apart; 50 and 100 steps give the same partition.

\subsection{Task-private deep l-stream}
\label{sec:ablation-private}
The private stream $\psi_{t}$ adds $L_{d}$ interventions per task. Table~\ref{tab:private_ablation} tests what it contributes. Shared stores every site in the shared intervention state, and private is the default. Dropped at test trains $\psi_{t}$ as usual but replaces it with the identity edit at test time, which measures how much task-specific signal the private stream carries. The private stream raises OP by 1.4 points on two seeds, because each task keeps capacity that no later join can overwrite. Dropping it at test time lowers OP by 3.6 points and raises BWT to 4.5, so the private stream holds task-specific signal that the shared state does not carry.
\begin{table}[t]
\centering
\footnotesize
\setlength{\tabcolsep}{2.5pt}
\caption{Ablation of the task-private deep l-stream on TRACE-8 with Llama-3.2-1B-Instruct, validation split, ablation configuration. Routed rows: mean of seeds 42 and 1234.}
\label{tab:private_ablation}
\begin{tabular}{l c c c c}
\toprule
Storage & $K$ & OP $\uparrow$ & BWT $\downarrow$ & BWT$_{\mathrm{in}}$ $\downarrow$ \\
\midrule
Shared, routed            & 7 & 45.27 & 0.02 & 0.18 \\
\textbf{Private, routed (CRAFT)} & 6 & \textbf{46.70} & 0.46 & 1.85 \\
Private, dropped at test  & 6 & 43.12 & 4.47 & -- \\
\bottomrule
\end{tabular}
\end{table}

\subsection{Normalizing the KL penalty}
Dividing the KL penalty by its running mean makes the anchor very strong whenever the KL is small, and tasks that join a shared intervention then barely learn (Table~\ref{tab:kl_norm}). CRAFT therefore uses the unnormalized penalty, so $\beta$ sets its weight directly.
\begin{table}[htbp]
\centering
\small
\setlength{\tabcolsep}{4pt}
\caption{Normalizing the KL penalty by its running mean, TRACE-8 validation split, Llama-3.2-1B-Instruct, seed 42, with $\delta = 1.5$ and at most three tasks per shared intervention. Both runs form the same four shared interventions. Py150 and ScienceQA join the shared intervention of MeetingBank.}
\label{tab:kl_norm}
\resizebox{\columnwidth}{!}{%
\begin{tabular}{l c c c c c}
\toprule
KL penalty & LA $\uparrow$ & OP $\uparrow$ & BWT $\downarrow$ & Py150 & ScienceQA \\
\midrule
$\beta\,\mathrm{KL} / \overline{\mathrm{KL}}$ & 32.60 & 32.51 & 0.14 & 8.75 & 18.60 \\
$\beta\,\mathrm{KL}$ (CRAFT) & \textbf{41.28} & \textbf{38.98} & 2.31 & \textbf{40.90} & \textbf{56.40} \\
\bottomrule
\end{tabular}}
\end{table}

\begin{table}[htbp]
\centering
\caption{Parameter complexity of continual-learning methods. $T$ is the number of tasks, $K \le T$ the number of shared interventions formed by routing, $L$ the number of layers, $h$ the hidden dimension, $r$ the LoRA and LoReFT rank, and $L_{d}$ the number of deep layers whose l-stream stays private to each task. With $K = T$, CRAFT costs as much as one intervention per task, and it costs less whenever $K < T$.}
\label{tab:param_complexity}
\small
\setlength{\tabcolsep}{4pt}
\begin{tabular}{ll}
\toprule
Method & Parameter complexity \\
\midrule
SeqLoRA & $\mathcal{O}(L h r)$ \\
LoRA per task & $\mathcal{O}(L h r T)$ \\
O-LoRA         & $\mathcal{O}(L h r T)$ \\
TreeLoRA       & $\mathcal{O}(L h r + T r)$ \\
Intervention per task & $\mathcal{O}(L h r T)$ \\
CRAFT (ours)   & $\mathcal{O}\bigl((K(2L - L_{d}) + T L_{d})\,h r\bigr)$ \\
\bottomrule
\end{tabular}
\end{table}
\begin{figure*}[t]
\centering
\includegraphics[width=0.8\textwidth]{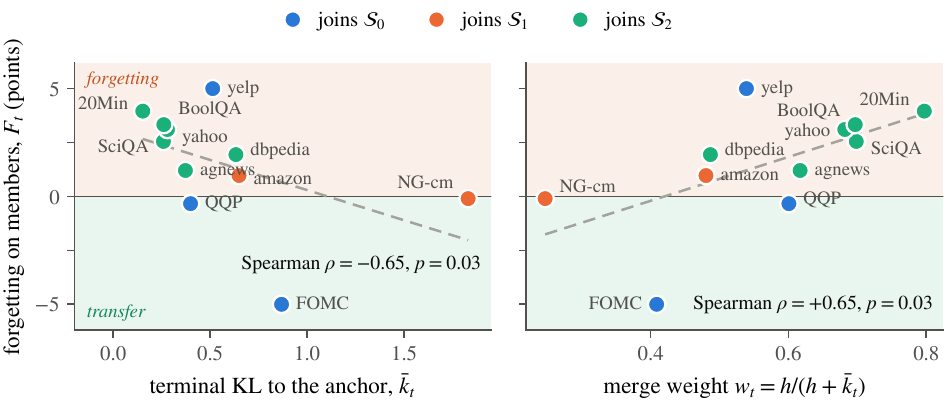}
\caption{Forgetting $F_t$ that each joining task causes on the earlier members of its shared intervention on the 15-task stream, against its terminal KL to the anchor (left) and its merge weight (right). Colors mark the shared intervention the task joins, and points below zero mark transfer. Tasks that drift further are merged with a smaller weight, so forgetting follows the merge weight rather than the KL.}
\label{fig:kl-dynamics}
\end{figure*}
\subsection{Empirical KL--forgetting relationship}
\label{sec:kl-forgetting-empirical}
\citet{shenfeld2026rl} report that forgetting grows with the KL between the pre- and post-finetuning output distributions on the new task. In CRAFT, this KL is measured before the merge, and the merge itself decides how much of the drift is kept. Figure~\ref{fig:kl-dynamics} relates the two on the 15-task stream for the 11 tasks that join an existing shared intervention, with $F_t$ the mean drop on the earlier members of that shared intervention when task $t$ is committed. Forgetting falls as the terminal KL rises (Spearman $\rho = -0.65$, $p = 0.03$) and rises with the merge weight ($\rho = 0.65$, $p = 0.03$). NumGLUE-cm ends with the largest KL, 1.83, is merged with a weight of 0.25 and leaves its members unchanged, while the largest losses, 2.6 to 5.0 points, come from tasks merged with weights of 0.54 to 0.80. The KL-weighted merge therefore turns a large drift into a small commit, and the forgetting that remains follows the share of the update that is committed, in line with the merge half-life sweep of Table~\ref{tab:components}. FOMC lies below zero because its commit raises C-STANCE from 51 to 56 points.
\begin{table*}[htbp]
\centering
\scriptsize
\setlength{\tabcolsep}{2.5pt}
\renewcommand{\arraystretch}{1.15}
\caption{Per-task evaluation matrix on the 15-task stream (Llama-3.2-1B-Instruct, seed 42). Rows are the most recently trained task; columns are the evaluation task. The diagonal shows performance right after training; entries below it show retention. Metrics as in ~\ref{:hyper_param}, in percent. Columns: CST = C-STANCE, MB = MeetingBank, Py = Py150, SQA = ScienceQA, NG = NumGLUE, 20M = 20Minuten, DBp = dbpedia, Amz = amazon, Yah = yahoo, AGN = agnews.}
\label{tab:eval_matrix_15task}
\begin{tabular}{l ccccccccccccccc}
\toprule
\multicolumn{1}{c}{\scriptsize Train $\setminus$ Test}
& CST & FOMC & MB & Py & SQA & NG-cm & NG-ds & 20M & DBp & Amz & Yah & AGN & Yelp & BoolQ & QQP \\
\midrule
T1  C-STANCE    & 51.0 &      &      &      &      &      &      &      &      &      &      &      &      &      &      \\
T2  FOMC        & 56.0 & 58.0 &      &      &      &      &      &      &      &      &      &      &      &      &      \\
T3  MeetingBank & 56.0 & 58.0 & 23.5 &      &      &      &      &      &      &      &      &      &      &      &      \\
T4  Py150       & 56.0 & 58.0 & 23.5 & 42.9 &      &      &      &      &      &      &      &      &      &      &      \\
T5  ScienceQA   & 56.0 & 58.0 & 23.5 & 40.3 & 48.0 &      &      &      &      &      &      &      &      &      &      \\
T6  NumGLUE-cm  & 56.0 & 58.0 & 23.6 & 40.3 & 48.0 & 18.5 &      &      &      &      &      &      &      &      &      \\
T7  NumGLUE-ds  & 56.0 & 58.0 & 23.6 & 40.3 & 48.0 & 18.5 & 49.0 &      &      &      &      &      &      &      &      \\
T8  20Minuten   & 56.0 & 58.0 & 23.6 & 34.4 & 46.0 & 18.5 & 49.0 & 41.0 &      &      &      &      &      &      &      \\
T9  dbpedia     & 56.0 & 58.0 & 23.6 & 31.7 & 43.0 & 18.5 & 49.0 & 40.9 & 79.0 &      &      &      &      &      &      \\
T10 amazon      & 56.0 & 58.0 & 22.9 & 31.7 & 43.0 & 17.3 & 49.0 & 40.9 & 79.0 & 45.0 &      &      &      &      &      \\
T11 yahoo       & 56.0 & 58.0 & 22.9 & 25.6 & 38.0 & 17.3 & 49.0 & 40.6 & 78.0 & 45.0 & 64.0 &      &      &      &      \\
T12 agnews      & 56.0 & 58.0 & 22.9 & 21.1 & 35.0 & 17.3 & 49.0 & 40.1 & 80.0 & 45.0 & 64.0 & 87.0 &      &      &      \\
T13 yelp        & 49.0 & 55.0 & 22.9 & 21.1 & 35.0 & 17.3 & 49.0 & 40.1 & 80.0 & 45.0 & 64.0 & 87.0 & 42.0 &      &      \\
T14 BoolQA      & 49.0 & 55.0 & 22.9 & 19.3 & 39.0 & 17.3 & 49.0 & 39.9 & 72.0 & 45.0 & 56.0 & 81.0 & 42.0 & 54.0 &      \\
T15 QQP         & 49.0 & 52.0 & 22.9 & 19.3 & 39.0 & 17.3 & 49.0 & 39.9 & 72.0 & 45.0 & 56.0 & 81.0 & 46.0 & 54.0 & 71.0 \\
\bottomrule
\end{tabular}
\end{table*}
\section{Trainable parameters and training cost}

\section{Conclusion}
\label{sec:conclusion}
This paper introduced CRAFT, a continual learning framework that keeps the backbone of a large language model frozen and learns each task through low-rank interventions on its hidden representations. Similar tasks share an intervention, while dissimilar tasks receive separate ones, and a small private part of each intervention stays with its own task. A single signal, the KL divergence between output distributions, guides the whole process: it routes each new task to a shared intervention or opens a new one, limits how far training moves from what that intervention already encodes, and sets how much of the trained state is merged back. 
\section*{Limitations}
CRAFT is designed for task-sequential continual learning, the protocol of TRACE and O-LoRA, in which tasks arrive one after another with known boundaries. This work studies it on text streams of up to fifteen tasks and on backbones of up to seven billion parameters. Streams of hundreds of tasks, larger backbones and multimodal models would test how the stored state grows and whether output-space routing transfers to other model families and inputs. Future work will adapt the routing threshold to each stream from the distances observed during training, and will extend CRAFT to federated continual learning, where clients exchange only interventions and KL routing can decide which clients' tasks share one. Since CRAFT edits representations rather than weights, it can also be combined with rehearsal and with other parameter-efficient adapters.

\bibliography{custom}
\
\section{Literature review}
\label{sec:litreview}
\subsection{Continual Learning for LLMs}
The TRACE benchmark~\citep{wang2023trace} established the current protocol for evaluating sequential task adaptation across classification, summarization, code, and reasoning. Existing approaches fall into three families that all act in parameter space. Regularization methods (EWC~\citep{kirkpatrick2017overcoming}, SI~\citep{zenke2017continual}) penalize changes to weights deemed important for prior tasks. Replay methods such as Self-Synthesized Rehearsal~\citep{huang2024mitigating} generate surrogate examples but still tune $\theta$ on them. Architecture-based methods isolate task knowledge into adapters: O-LoRA~\citep{wang2023orthogonal} learns each task in a subspace orthogonal to past tasks, TreeLoRA~\citep{qian2025tree} organizes layer-wise LoRA adapters by gradient similarity, and Mixture-of-LoRA variants (MoELoRA~\citep{liu2023moelora}, MoRAL~\citep{yang2024moral}, D-MoLE~\citep{ge2025dynamic}) maintain adapter pools and pair them with a learned gating network trained jointly with the experts. CRAFT is structurally similar in that it also maintains a small pool of similar-task intervention states, but task assignment during training uses no learned router and no gating parameters. For task-agnostic routing at inference, CRAFT reads the representation after each learned shared intervention with one linear classifier per shared intervention, refit from stored features at every commit, so no classifier is trained sequentially. Its closest framing is therefore a representation-level mixture-of-experts with output-space routing.

Across all three families, every gradient step competes for weights that encode prior knowledge, so the stability--plasticity tradeoff is intrinsic. Functionally, however, CRAFT is orthogonal to all three families: by lifting adaptation into representation space, the gradient steps for task $t$ never touch the weights that encode any prior task, so the stability--plasticity tradeoff is recast from a parameter-space tug-of-war into a question of how to organize interventions.

\paragraph{Recent low-rank and representation-space continual learning:}
GainLoRA~\citep{liang2025gainlora} adds a LoRA branch per task and learns gates that limit a new branch's influence on old tasks; CRAFT weights its commit by an output-space KL and trains no gate. ProCL~\citep{le2026procl} retrieves LoRA memory slots by input-conditioned attention, and AM-LoRA~\citep{liu2024amlora} mixes task-specific LoRAs with attention; both route in parameter space with learned modules, while CRAFT routes by output divergence during training. InfLoRA~\citep{liang2024inflora}, O-LoRA~\citep{wang2023orthogonal} and OSFT~\citep{nayak2025osft} restrict weight updates to low-interference subspaces; CRAFT leaves the weights frozen and constrains outputs instead. The closest work is CoRe~\citep{luo2026core}, which applies task-specific representation interventions to continual learning in vision transformers; CRAFT shares interventions across tasks through output-space routing and a KL anchor, and targets language models.

\subsection{Representation Fine-Tuning}
LoReFT, introduced by \citet{wu2404reft}, is an orthonormal-subspace edit on hidden states with $\theta$ frozen. With fewer trainable intervention parameters than LoRA, LoReFT matches or exceeds parameter-space PEFTs on commonsense reasoning, instruction following, and NLU. Two properties matter for continual learning: $\theta$ being frozen makes weight-level interference structurally impossible (only $\phiparam^{(t)}$ can be overwritten), and the intervention has an explicit target $\Wmat h + \bvec$ inside the rank-$r$ subspace rather than an implicit one mediated by full-parameter loss minimization. Prior representation fine-tuning work has been confined to single-task settings; its extension to a continual stream of heterogeneous tasks, with routing, anchoring, and merging at the intervention level, is the gap we address.

\subsection{KL Regularization}
Combining a fine-tuning loss with a KL penalty against an anchor distribution is well established in RLHF~\citep{stiennon2020learning, ouyang2022training, korbak2022rl, vieillard2020leverage}. \citet{chen2025retaining} and \citet{shenfeld2026rl} show that on-policy RL fine-tuning forgets less than SFT at matched new-task performance; \citet{shenfeld2026rl} identifies the mechanism (forgetting is predicted by KL on the new task, and policy gradient is implicitly biased toward KL-minimal solutions). These works operate in parameter space against the base model. We bring the principle into representation space and use the current merged state of the shared intervention as the anchor; the KL penalty restrains drift during training, and the same KL sets how much of the trained state replaces the anchor at commit.

\section{Experimental details}
\label{:hyper_param}
\subsection{Metrics}
Following \citet{wang2023trace}, $a_{j,t}$ denotes the score on task $t$ after training up to task $j$. Classification tasks (C-STANCE, FOMC, ScienceQA and the added classification tasks) use accuracy, MeetingBank uses ROUGE-L, 20Minuten uses SARI, NumGLUE-cm and NumGLUE-ds use exact match, and Py150 uses edit-distance similarity. Overall performance (OP) is the mean score on all $T$ tasks after the last task is trained. Backward transfer (BWT) is the mean drop of each task from the end of its own training to the end of the stream, $\frac{1}{T}\sum_{t}a_{t,t}-a_{T,t})$, computed as in the TreeLoRA implementation: a task that improves counts as zero, and a positive value means forgetting. Learning accuracy (LA), the mean of $a_{t,t}$, measures how well each task is learned at the moment it is trained. $\mathrm{BWT}_{\mathrm{in}}$ restricts BWT to the tasks that share their intervention with a later task, the only tasks that CRAFT can change after their own training.
\subsection{Benchmarks}
We evaluate on a 15-task sequence consisting of the TRACE benchmark ~\citep{wang2023trace} and seven additional classification tasks.
\paragraph{TRACE benchmark:}
C-STANCE is a Chinese stance detection task.  FOMC focuses on financial-domain classification based on central bank communications.  MeetingBank is a long-form summarization task that requires generating concise summaries from meeting transcripts.  20Minuten is a German text simplification task on news articles.  Py150 is a Python code completion task.  ScienceQA is a reasoning task that involves answering science questions requiring multi-step understanding.  NumGLUE-cm evaluates numerical reasoning in controlled settings, while NumGLUE-ds focuses on more diverse and realistic numerical reasoning scenarios.

\paragraph{Additional classification tasks:} AG News is a topic classification task over news articles ~\citep{wang2023orthogonal}. Amazon and Yelp are sentiment classification tasks based on customer reviews. DBpedia and Yahoo are large-scale topic classification datasets with diverse categories. BoolQ is a binary question answering task requiring yes/no decisions. QQP (Quora Question Pairs) is a paraphrase detection task that determines whether two questions have the same meaning.

\subsection{Backbones}
The main comparison uses three foundation large language models. These models cover a range of architectures and parameter scales, enabling evaluation across different capacities and design choices. The key characteristics of each model are summarized below:
\begin{itemize}
    \item Llama-3.2-1B-Instruct: A lightweight instruction-tuned model with 1 billion parameters, optimized for efficient deployment. It has 16 Transformer layers, 2048-dimensional hidden states, and 32 attention heads with 8 key-value heads, and is trained on curated instruction-following data.
    \item Gemma-2B-it: A compact open model from Google with 2 billion parameters. It uses multi-query attention and consists of 18 Transformer layers. The model is designed for efficient reasoning and instruction-following, with built-in safety and alignment constraints.

    \item Llama-2-7B-Chat: A decoder-only Transformer with 7 billion parameters, trained on approximately 2 trillion tokens. It incorporates supervised fine-tuning and alignment techniques, with 32 Transformer layers, 4096-dimensional hidden states, and 32 attention heads.

\end{itemize}

\subsection{Baselines}
\label{:baselines}
To rigorously evaluate our approach, we benchmark against several established continual learning (CL) and parameter-efficient tuning methods:

\begin{itemize}
    
    \item SeqLoRA trains a single LoRA adapter on the tasks one after another, with no mechanism against forgetting.

    \item GEM \citep{lopez2017gradient} An episodic memory-based strategy that caches a small subset of historical task examples. During training, it projects current gradients to ensure they do not interfere with the gradients of previously learned tasks.
    
    \item EWC \citep{kirkpatrick2017overcoming} A well-known regularization technique that utilizes the Fisher Information Matrix to estimate the importance of specific network weights, penalizing significant updates to parameters deemed critical for earlier tasks.
    
    \item L2P \citep{wang2022learning} This method leverages a prompt-pooling technique to facilitate continual learning, creating a unified repository of prompts accessible across multiple sequential tasks. By associating each prompt with a unique key vector, the system dynamically retrieves and applies these prompts to the final multi-head self-attention (MSA) layer, instructing the network via Prompt Tuning
    
    \item DualPrompt \citep{wang2022bDualPrompt} Building upon pooled prompting, this approach separates instructions into task-invariant and task-specific expert prompts. These are applied across various attention layers using a Prefix-Tuning methodology to optimize knowledge transfer.

    \item HiDeLoRA \citep{wang2025hide} An evolution of the hierarchical decomposition strategy that applies the concept to low-rank adapters rather than prompts, facilitating highly efficient parameter sharing across a unified CL framework.
    
    \item O-LoRA \citep{wang2023orthogonal} This approach restricts the learning of new tasks to distinct, orthogonal low-rank subspaces, inherently avoiding parameter overlap and mitigating catastrophic interference between tasks.

\end{itemize}

\subsection{Hyperparameters and implementation details}
Table~\ref{tab:_craft_hyper_all} lists the settings of every reported CRAFT run, taken from the run scripts. Shared interventions have no size limit in the main runs, unless stated. 
\paragraph{Setting CRAFT for a new model.} Every routing setting is shared by all backbones. Per backbone, only the learning rate is chosen on the validation split, and Gemma-2B-it uses a weaker anchor ($\beta = 0.2$) and a longer merge half-life ($h = 1.0$). For a new model, keep the routing defaults and choose the learning rate by validation OP. Any $\beta$ from 0.1 to 1.0 keeps validation BWT below 0.4 and changes OP by at most 2 points (Table~\ref{tab:sensitivity}), so $\beta$ needs no tuning.

\paragraph{Intervention placement.} Interventions sit on the first and last $t_{pos} = 15$ tokens of every input at all layers with rank $r = 8$. On the three long-input tasks, MeetingBank, 20Minuten and Py150, the l-stream covers the last 30 positions on Llama-3.2-1B-Instruct and Gemma-2B-it. The deepest 38\% of layers keep their l-stream private to each task.

\paragraph{Hyperparameters.}
CRAFT is stable over wide ranges of each training hyperparameter ( Table~\ref{tab:sensitivity}). Anchor strengths from 0.1 to 1.0 keep BWT between 0.19 and 0.40 and OP between 45.9 and 47.9, because even a weak anchor keeps a joining task near its shared state, and positions per stream from 9 to 21 and ranks 4 and 8 stay within one point of the reference. The routing threshold $\delta$ sets how readily tasks share: OP is highest at $\delta = 0.7$ and drops by about four points once $\delta \ge 1.0$ places more tasks together, while BWT stays below 0.8 ( Table~\ref{tab:delta_nocap}). \\

\paragraph{Optimization.} Each task trains for its full epoch budget of 5, 4, 7, 5, 4, 5, 5 and 7 epochs on the eight TRACE tasks and 5 epochs on each added task of the 15-task stream, with 5{,}000 training examples per task on TRACE and 500 per task on the 15-task stream. Optimization uses AdamW with no weight decay and no dropout, a linear schedule with 5\% warm-up, batch size 4 and no gradient accumulation. The KL anchor is computed on the positions that predict answer tokens, and the merge weight uses the KL on the training data of the joining task, averaged over 50 batches.

\paragraph{Ablation configuration.} The component ablations and the sensitivity study run on the validation split of TRACE-8 with Llama-3.2-1B-Instruct, with a capacity of two tasks per shared intervention. The reference run and every anchor and merge variant share one partition of the tasks, so each row of Table~\ref{tab:components} changes one component on the same partition. Table~\ref{tab:sensitivity} reports the partition size of each setting.

\begin{table*}[ht]
\centering
\caption{Hyperparameter settings of the reported CRAFT runs on TRACE-8, taken from the run scripts.}
\label{tab:_craft_hyper_all}
\footnotesize
\resizebox{\textwidth}{!}{%
\begin{tabular}{lcccc}
\toprule
\textbf{Hyperparameter} & \textbf{Llama-3.2-1B-Instruct} & \textbf{Gemma-2B-it} & \textbf{Llama-2-7B-Chat} & \textbf{R1-Distill-Qwen-1.5B} \\
\midrule
Epochs              & 5,4,7,5,4,5,5,7 & 5,4,7,5,4,5,5,7 & 5,4,7,5,4,5,5,7 & 5,4,7,5,4,5,5,7 \\
Examples per task   & 5{,}000 & 5{,}000 & 5{,}000 & 5{,}000 \\
Learning rate       & $2\times10^{-4}$ & $2\times10^{-4}$ & $10^{-4}$ & $2\times10^{-4}$ \\
Anchor $\beta$      & 0.3 & 0.2 & 0.3 & 0.3 \\
Merge half-life $h$ & 0.6 & 1.0 & 0.6 & 0.6 \\
Max length          & 2048 & 1024 & 1024 & 2048 \\
Positions           & f15+l15 & f15+l15 & f15+l15 & f15+l15 \\
\midrule
Rank $r$                         & \multicolumn{4}{c}{8} \\
Layers                           & \multicolumn{4}{c}{all} \\
Private share $\rho$             & \multicolumn{4}{c}{0.38} \\
Routing direction                & \multicolumn{4}{c}{member-side} \\
Threshold $\delta$               & \multicolumn{4}{c}{0.7} \\
Shared-intervention capacity $C$ & \multicolumn{4}{c}{unlimited} \\
Floors $\varepsilon_{\mathrm{b}}, \varepsilon_{\mathrm{sig}}$ & \multicolumn{4}{c}{0.05, 0.01} \\
Warm-up $S_{\mathrm{wu}}$        & \multicolumn{4}{c}{100} \\
Probe size, top-$k$, temperature & \multicolumn{4}{c}{16, 512, 2.0} \\
Merge-KL data, batches           & \multicolumn{4}{c}{training data, 50} \\
Task-agnostic routing $n_{h}$, $m$          & \multicolumn{4}{c}{64, 16} \\
Task-agnostic routing $\ell_2$, iterations  & \multicolumn{4}{c}{$10^{-2}$, 200} \\
\midrule
Dropout             & \multicolumn{4}{c}{0.00} \\
Optimizer           & \multicolumn{4}{c}{AdamW} \\
Weight decay        & \multicolumn{4}{c}{0.00} \\
Warm-up ratio, schedule & \multicolumn{4}{c}{0.05, linear} \\
Batch size          & \multicolumn{4}{c}{4} \\
Gradient accum.     & \multicolumn{4}{c}{1} \\
Eval batch size     & \multicolumn{4}{c}{4} \\
\bottomrule
\end{tabular}}
\end{table*}
\section{Implementation details}
\subsection{One CRAFT step}
\begin{algorithm}[t]
\caption{One CRAFT step for task $t$.}
\label{alg:craft}
\begin{algorithmic}[1]
\Require $\mathcal{D}_t$; shared intervention states $\{\phiparam_{\mathcal{S}_k}\}$; stored probes $\{\mathcal{P}_s\}$
\State $\mathcal{P}_t \gets$ $n_p$ training prompts of $\mathcal{D}_t$
\If{no shared intervention exists}
  \State $k^\star \gets$ new shared intervention
\Else
  \State $\widetilde{\phiparam}^{(t)} \gets$ $S_{\mathrm{wu}}$ SFT steps from the fixed-seed initialization
  \State $d(t,k) \gets$ Eq.~\eqref{eq:route_dist} for every shared intervention
  \State $k^\star \gets \argmin_k d(t,k)$
  \If{$d(t,k^\star) > \delta$ \textbf{or} $D_{KS_{k^\star}} < \varepsilon_{\mathrm{sig}}$}
    \State $k^\star \gets$ new shared intervention
  \EndIf
\EndIf
\If{$k^\star$ is not new}
  \State $\phiparam^{\mathrm{anchor}} \gets$ shared sites of $\phiparam_{\mathcal{S}_{k^\star}}$, private sites at initialization; $\phiparam^{(t)} \gets \phiparam^{\mathrm{anchor}}$
  \State train on Eq.~\eqref{eq:kl_loss} for the full budget
\EndIf
\If{$k^\star$ is new}
  \State train $\phiparam^{(t)}$ from the fixed-seed initialization on $\mathcal{L}_\tau$ for the full budget
  \State $\phiparam_{\mathcal{S}_{k^\star}} \gets \phiparam^{(t)}_{\text{shared}}$
\Else
  \State $w_t \gets h / (h + \bar{k}_t)$; $\phiparam_{\mathcal{S}_{k^\star}} \gets w_t\,\phiparam^{(t)}_{\text{shared}} + (1 - w_t)\,\phiparam_{\mathcal{S}_{k^\star}}$
\EndIf
\State store $\psi_t$, $\mathcal{P}_t$, $n_{h}$ prompts of $t$, and $M[t] \gets k^\star$ \State recompute shared intervention $k^\star$'s task-agnostic routing features on all stored prompts and the other shared interventions' features on the new prompts; refit every $\omega_k$ from zero
\end{algorithmic}
\end{algorithm}
\subsection{Routing divergences}
For intervention states $A$ and $B$, let $z_A(x,j)$ be the logits at position $j$ of probe sequence $x$. Let $V_A$ and $V_B$ be the top-$k$ tokens of each, with $k = 512$, and let $q_{A,v}$ be the softmax of $z_A / T$ over $V_A$ at token $v$, with $T = 2$. At each position,
\begin{equation}
  d_{AB} = \tfrac{1}{2}\!\! \sum_{v \in V_A \cap V_B} (q_{A,v} - q_{B,v}) \log \tfrac{q_{A,v}}{q_{B,v}},
  \label{eq:route_div}
\end{equation}
and $D(A, B; \mathcal{P})$ is the mean of $d_{AB}$ over all $x \in \mathcal{P}$ and all $j \in \mathcal{J}(x)$, the positions that predict answer tokens. Then $D_{K} = D(\varnothing, \widetilde{\phiparam}^{(t)})$, $D_{S_k} = D(\varnothing, \phiparam_{\mathcal{S}_k})$ and $D_{KS_k} = D(\widetilde{\phiparam}^{(t)}, \phiparam_{\mathcal{S}_k})$, where $\varnothing$ is the backbone with no intervention. The member-side distance evaluates them on $\mathcal{P}_s$ for each member $s$, and the no-signal floor evaluates $D_{KS_{k^\star}}$ on $\mathcal{P}_t$. Each term is non-negative, the divergence has no upper bound, and tokens ranked highly by only one state do not contribute.

\subsection{Rotation matrix}
The projection $\Rmat$ is constrained to the Stiefel manifold through pyreft's orthogonal parameterization. pyreft stores the realized matrix $\Rmat$ and, on loading, writes it into the fixed base of a freshly built orthogonal parametrization, so a loaded state reproduces $\Rmat$ exactly. The merge of Eq.~\eqref{eq:kl_merge} interpolates these realized matrices. With $\Delta = \Rmat_{\mathrm{new}} - \Rmat_{\mathrm{old}}$ and weight $w$, the committed matrix $\Rmat_{\mathrm{c}}$ satisfies
\begin{equation}
  \Rmat_{\mathrm{c}}\Rmat_{\mathrm{c}}^{\top} = \bm{I} - w(1-w)\,\Delta\Delta^{\top},
  \label{eq:R_merge}
\end{equation}
so a merge keeps $\Rmat_{\mathrm{c}}$ on the Stiefel manifold only up to the term $w(1-w)\,\Delta\Delta^{\top}$. Eq.~\eqref{eq:R_merge} holds exactly when both matrices are orthonormal, as in the first merge into a shared intervention; later merges start from an already merged state, so the deviation accumulates. A state that has never been merged stays orthonormal to numerical precision ($\|\Rmat\Rmat^{\top} - \bm{I}\|_F \approx 10^{-6}$). Over all stored shared intervention states, the largest deviation $\|\Rmat\Rmat^{\top} - \bm{I}\|_F$ is 0.46 on the 15-task stream, where a shared intervention absorbs up to six merges, and 0.65 and 1.21 on TRACE-8 with Gemma-2B-it and Llama-3.2-1B-Instruct, where it absorbs at most two and three merges, against $\|\bm{I}_r\|_F = \sqrt{8} \approx 2.83$. The deviation therefore depends on how far a joining task rotates the subspace more than on the number of merges. The smallest merge weight in these runs is $w_{\min} = 0.25$. All reported scores are measured with these merged states. The edit of Eq.~\eqref{eq:loreft} is well defined for any $\Rmat$; only its reading as an orthogonal projection is approximate.

\section{Task-agnostic routing}
\label{app:litar}
\paragraph{Features.} For shared intervention $k$, the backbone runs once with $\phiparam_{\mathcal{S}_k}$ loaded and the private sites set to the identity edit, so only the shared state acts. The hidden states of prompt $x$ are averaged over its last $t_{pos} = 15$ prompt tokens at every layer and concatenated into $\bm{z}_k(x)$. Every hidden state has passed through the LoReFT edits of the shared intervention at the edited positions, and every later layer sees the edited input, so the features carry what the shared intervention learned.

\paragraph{Classifiers.} For each shared intervention, the features of $n_h = 64$ training prompts of every task form a bank, and one multinomial logistic regression $\omega_k$ is fitted from zero on standardized features with cross-entropy and an $\ell_2$ penalty of $10^{-2}$, for at most 200 L-BFGS steps. Because the backbone never changes, a bank changes only when a task is merged into its shared intervention. CRAFT then recomputes that bank from the stored prompts and refits its classifier, so every refit is deterministic and no classifier is trained sequentially.

\paragraph{Decision and cost.} Eq.~\eqref{eq:task_head_decision} averages the $K$ classifiers and the first $m = 16$ prompts of a test task, and $m = 1$ routes prompts that arrive interleaved. At test time, task-agnostic routing needs $K$ forward passes without generation per prompt, one per shared state, and $K$ products with a $T \times Ld$ weight matrix. A bank takes about 4.2\,MB per task and shared intervention on Llama-3.2-1B-Instruct. On the Llama-3.2-1B-Instruct run of Table~\ref{tab:routing_ablation}, task-agnostic routing names the correct task for all 32 blocks of 16 test prompts and for 510 of 512 single prompts, and both misses still select the correct shared intervention.

% \begin{figure}[t]
% \centering
% \fbox{\parbox[c][3.2cm][c]{0.92\linewidth}{\centering\textcolor{red}{Router}}}
% \caption{Task-agnostic routing as tasks arrive on TRACE-8 with Llama-3.2-1B-Instruct. Left: task-identification accuracy after each task. Right: OP with task-agnostic routing against task-aware lookup.}
% \label{fig:litar}
% \end{figure}

\subsection{Trainable parameters}
CRAFT is compared with CL baselines that store per-task adapters at rank $r=8$. LoRA injects rank-$r$ updates into selected weight projections, paying a per-layer cost at every adapted projection. LoReFT instead edits the residual stream at a small set of token positions, with parameters shared across positions within a layer, so the cost is paid once per intervened layer. One intervention per task therefore costs about as much as one LoRA adapter on $q,v$ per task, and sharing brings CRAFT to about 40\% of either. CRAFT shares every site except the deep l-stream across the shared interventions discovered by routing, so its footprint is $K$ shared states plus $T$ private blocks of $L_{d}$ interventions each.

LoRA-style methods that allocate a fresh adapter per task scale linearly in $T$ in Table \ref{tab:param_complexity}. CRAFT scales mainly with the number of shared interventions $K$ discovered by routing. Only the $L_{d}$ private interventions per task grow with $T$. TreeLoRA achieves the smallest asymptotic footprint by sharing a single backbone adapter across all tasks with only a small $\mathcal{O}(T r)$ per-task term; CRAFT trades this against per-shared-intervention isolation in representation space.
\subsection{Training cost}
Training the TRACE-8 stream on Llama-3.2-1B-Instruct takes 5.9 hours on one A100 with CRAFT, against 5.2 hours when each task stores its own intervention. The extra time goes to the warm-up and the routing probes. A single intervention trained with a KL anchor also takes 5.9 hours, because every task after the first trains against an anchor, while CRAFT pays for the anchor's frozen forward pass only on tasks that join a shared intervention.
\section{Notation}
\label{:notation}
\begin{table}[htbp]
\centering
\caption{Notation used throughout the paper.}
\small
\setlength{\tabcolsep}{4pt}
\renewcommand{\arraystretch}{1.15}
\begin{tabular}{@{}l p{0.56\columnwidth}@{}}
\toprule
\textbf{Symbol} & \textbf{Meaning} \\
\midrule
$\theta$            & Frozen backbone weights \\
$L$                 & Number of transformer layers \\
$\hvec^{(l)}$       & Hidden state at layer $l$ \\
$\pi_{0}$           & Backbone output distribution with no intervention \\
\midrule
$\phiparam = \{\Rmat, \Wmat, \bvec\}$ & LoReFT intervention parameters (rank $r$) \\
$\Phi(\hvec; \phiparam)$ & Intervention function (Eq.~\eqref{eq:loreft}) \\
$\pi_{\phiparam}$   & Output distribution induced by intervention $\phiparam$ \\
$t_{pos}$           & Number of intervention positions per stream (f / l) \\
\midrule
$T$, $t$, $\Dt$ & Task stream length; task $t$ and its data \\
$K$, $\mathcal{S}_{k}$ & Number of similar-task shared interventions; shared intervention $k$ \\
$\phiparam_{\mathcal{S}_{k}}$ & State of shared intervention $k$ \\
$\phiparam^{(t)}$   & Intervention parameters trained for task $t$ \\
$\phiparam^{\mathrm{anchor}}$ & Frozen snapshot of $\phiparam_{\mathcal{S}_{k}}$ before training $t$ \\
$\widetilde{\phiparam}^{(t)}$ & Warmup-end intervention for task $t$ \\
$\psi_{t}$ & Private deep l-stream of task $t$ \\
$L_{d}$, $\rho$ & Private layers; their share of $L$ \\
$\mathcal{P}_{s}$, $n_{p}$ & Routing probe of task $s$; its size \\
$\varepsilon_{\mathrm{b}}$, $\varepsilon_{\mathrm{sig}}$ & Denominator floor; no-signal floor \\
$C$ & Shared-intervention capacity \\
$h$, $w_{t}$ & Merge half-life; merge weight \\
$p_{\omega}$, $n_{h}$, $m$ & Task head; stored and test prompts \\
\midrule
$\KLdiv(p \,\|\, q)$ & Forward KL divergence \\
$d(t, k)$           & Routing distance, normalized symmetric KL (Eq.~\eqref{eq:route_dist}) \\
$\delta$, $S_{\mathrm{wu}}$ & Routing threshold; warm-up steps \\
$\beta$             & KL anchor coefficient in $\mathcal{L}_{\mathrm{total}}$ \\
\midrule
$a_{j, t}$          & Score on task $t$ after training up to task $j$ \\
$\mathrm{OP}$, $\mathrm{BWT}$ & Overall performance ($\uparrow$); backward transfer ($\downarrow$) \\
$p_{\omega_k}$, $n_{h}$, $m$ & Task-agnostic routing classifier of shared intervention $k$; stored and test prompts \\
\bottomrule
\end{tabular}
\label{tab:notation}
\end{table}
\end{document}